\documentclass{article}

\usepackage{arxiv}

\usepackage[utf8]{inputenc} 
\usepackage[T1]{fontenc}    
\usepackage{hyperref}       
\usepackage{url}            
\usepackage{booktabs}       
\usepackage{amsfonts}       
\usepackage{nicefrac}       
\usepackage{microtype}      
\usepackage{lipsum}
\usepackage{graphicx}

\usepackage{amssymb}
\usepackage{amsmath}
\usepackage{algorithm}
\usepackage{algorithmic}

\renewcommand{\today}{}

\title{MicroFlow: An Efficient Rust-Based Inference Engine for TinyML}

\author{
 Matteo Carnelos\\
  University of Padua, Italy\\
  Grepit AB, Sweden\\
  \texttt{matteo.carnelos@studenti.unipd.it} \\
   \And
 Francesco Pasti \\
  University of Padua, Italy\\
  \texttt{pastifranc@dei.unipd.it} \\
  \And
 Nicola Bellotto \\
  University of Padua, Italy\\
  \texttt{nbellotto@dei.unipd.it} \\
}

\date{}

\begin{document}
\maketitle

\begin{abstract}
In recent years, there has been a significant interest in developing machine learning algorithms on embedded systems. This is particularly relevant for bare metal devices in Internet of Things, Robotics, and Industrial applications that face limited memory, processing power, and storage, and which require extreme robustness.
To address these constraints, we present MicroFlow, an open-source TinyML framework for the deployment of Neural Networks~(NNs) on embedded systems using the Rust programming language. The compiler-based inference engine of MicroFlow, coupled with Rust's memory safety, makes it suitable for TinyML applications in critical environments.
The proposed framework enables the successful deployment of NNs on highly resource-constrained devices, including bare-metal 8-bit microcontrollers with only 2kB of RAM.
Furthermore, MicroFlow is able to use less Flash and RAM memory than other state-of-the-art solutions for deploying NN reference models (i.e. wake-word and person detection), achieving equally accurate but faster inference compared to existing engines on medium-size NNs, and similar performance on bigger ones.
The experimental results prove the efficiency and suitability of MicroFlow for the deployment of TinyML models in critical environments where resources are particularly limited.
\end{abstract}

\section{Introduction}
\label{sec:introduction}

TinyML is a field of Machine Learning~(ML) that focuses on small and low-power embedded devices~\cite{tinyml-book}.
The aim of TinyML is to enable these devices to perform intelligent tasks without relying on cloud-based servers or high-performance computing systems.
The field has gained popularity in recent years due to the increasing demand for smart devices capable of performing intelligent real-time tasks without the need of cloud resources, which are usually power consuming and potentially unsafe regarding data security and privacy.
The increasing emergence of Internet of Things~(IoT) devices in domestic and industrial environments also contributed significantly to the field, making it possible to achieve even more integrated applications~\cite{dutta2021tinyml}.

A significant advantage of TinyML is its ability to operate on low-power devices, making it ideal for resource-constrained environments.
That means devices can operate on batteries or solar power, enabling access to technology in areas with limited infrastructure.
As of 2022, the global Micro-Controller Unit~(MCU) market was valued at USD~25.48~billion and is expected to expand at a Compound Annual Growth Rate~(CAGR) of $11.2\%$ from 2023 to 2030~\cite{978-1-68038-141-2}.
With such a large market, combined with the rise in popularity of AI applications~\cite{2022-ai-index-report}, the potential for TinyML-powered devices is particularly high, with a lot of big companies investing in this research field~\cite{tinyml-comes-to-embedded-world-2023}.
Moreover, while traditional ML models require significant computing power and hardware resources, which can be expensive, TinyML models can run on low-cost MCU, making this technology even more accessible~\cite{9166461}.
These feature are particularly important in developing countries~\cite{9682107} where access to electricity and computing resources can be a significant issue.

One of the major challenges when it comes to developing applications for embedded devices is their limited computational power.
Traditional ML models and inference engines are often not suitable for IoT applications, since they require a significant amount of computing and storage resources.
TinyML solutions instead run on MCUs with limited memory and processing capabilities, but this limits the deployment of complex NN architectures.
Therefore, both inference engines and NN models need to be optimized to run on resource-constrained hardware.
The lack of standardization in the TinyML ecosystem can pose further challenges for developers.
With the wide range of hardware platforms available, it can be difficult to ensure compatibility and interoperability between different components.

Some popular TinyML frameworks already exist, such as TensorFlow Lite for Microcontrollers~(TFLM) and others~\cite{MLSYS2021_d2ddea18,dutta2021tinyml}.
However, although these engines are widely used and well developed, they share some criticalities.
In particular, despite their optimizations, most of them require a significant amount of memory to run, which can be a challenge for small embedded devices.
Moreover, they are mostly available for 32-bit MCU and, in some cases, for specific architectures (e.g. Arm) or as proprietary software, which can lead to compatibility and interoperability issues.
Lastly, but perhaps more importantly, is the fact that they are typically written in C and C++.
Although these languages are the standard for embedded software, they are generally considered memory unsafe~\cite{WhiteHouse2024}.
This is a problem for TinyML applications in critical environments, where memory-related bugs and vulnerabilities are not acceptable.

Important but still unsolved research questions in this domain include understanding how to implement efficient TinyML applications for highly constrained embedded systems, investigating open-source solutions for both high- and low-end MCU architectures, and, most importantly, determining how to achieve memory-safe implementations.
To address these challenges and overcome the current limitations, this paper presents \emph{MicroFlow}, a new lightweight and inherently safe TinyML inference engine written in Rust, which is particularly designed for robustness and efficiency.
To achieve these goals, state-of-the-art approaches combined with newly proposed techniques have been used.
In particular, MicroFlow introduces three major contributions:
\begin{itemize}

\item {\bf Rust compiler for TinyML} --
MicroFlow has been developed using the Rust programming language, which provides inherent memory safety through its language features and guarantees. This means that MicroFlow is designed to prevent common memory-related errors such as null pointer dereferences, buffer overflows, and data races. By leveraging the power of Rust, MicroFlow is able to achieve a higher level of reliability and security in its memory management compared to existing C/C++ solutions.

\item {\bf Efficient memory allocation} --
In MicroFlow, the memory needed by the inference process is statically allocated, i.e. completely determined at compile-time and allocated before program execution. It is also automatically deallocated by the stack, avoiding manual memory management by the programmer.
This guarantees safe and efficient memory usage according to the model's requirements. 
Moreover, a new page-based method for memory access, where only subsets of the NN model are loaded into RAM for sequential processing, enables the inference on highly resource-constrained devices, including 8-bit MCUs.

\item {\bf Modular and open-source implementation} -- Microflow's code is completely open-source and freely available on GitHub\footnote{\url{https://github.com/matteocarnelos/microflow-rs}}. The repository includes the Rust source code and its documentation to facilitate reproducibility and further development by the Embedded Systems and IoT communities. Moreover, detailed experimental results are provided by benchmarking against the state-of-the-art in TinyML (i.e. TFLM).

\end{itemize}

The remainder of the paper is as follows. Sec.~\ref{sec:related-work} reviews the current state-of-the-art and provides some background information about the topic, identifying also gaps in the literature that MicroFlow seeks to address.
Sec.~\ref{sec:design} explains the design and implementation of the system with its main components.
Sec.~\ref{sec:memory_management} and Sec.~\ref{sec:operators} delve into MicroFlow's memory management and functionalities, respectively.
Section~\ref{sec:evaluation} presents the experimental evaluation, with a comprehensive analysis of the system's performance, effectiveness, and efficiency.
Finally, Sec.~\ref{sec:conclusions} summarises the paper and explores potential areas for future work.

\section{Related Work}
\label{sec:related-work}

TinyML is a relatively recent research area, the main concepts of which are therefore introduced in this section. An overview of programming languages and frameworks used for TinyML applications will follow, identifying gaps in the current state-of-the-art and motivating the solutions proposed by MicroFlow.

\subsection{TinyML Concepts}
TinyML refers to the deployment of ML models on small and resource-constrained embedded devices, enabling real-time inference without relying on powerful computers or cloud servers. This emerging research area has the potential to significantly enhance energy efficiency and reduce the latency of AI algorithms while guaranteeing privacy across a wide range of applications~\cite{lin2023embedded}.
As an example, one well-known and widely used TinyML applications is \emph{wake-word} detection, also called \emph{keyword spotting} or \emph{hotword}~\cite{6854370},  commonly implemented in voice assistants and smart speakers developed by companies such as Amazon, Apple, Google, and others.
This application involves training a NN to identify a specific sound or phrase, such as ``Hey Siri'' or ``OK Google'', that triggers a device to begin listening for user commands.
The trained model is then compressed and deployed to edge devices enabling local always-on inference greatly improving latency and privacy.
Other TinyML applications include activity recognition~\cite{labrador2013human}, object detection~\cite{9463524}, predictive maintenance~\cite{9837510}, environmental monitoring~\cite{9700573}, and many more.

A typical ML application involves two main phases: training and inference.
While the training phase is often the most computationally intensive~\cite{justus2018predicting}, TinyML research frequently explores methods for sparse model updates~\cite{lin2022device, pavan2024tybox} to enable on-device incremental learning or fine-tuning~\cite{ravaglia2021tinyml, pasti2024latent}.
Full model training, however, is usually performed on host systems equipped with high-performance computing resources like Graphics Processing Units (GPUs).

ML models require a considerable amount of memory storage and consume a substantial amount of CPU cycles also during the inference phase. To overcome these issues, the tinyML literature explores techniques such as \emph{quantization}, \emph{pruning}, \emph{knowledge distillation} and \emph{efficient architectures} to reduce memory and computational requirements. Quantization converts the floating-point values of the model parameters to fixed-point values with lower precision, typically 8-bit integers, significantly reducing storage and computation requirements~\cite{jacob2018quantization}. Pruning, on the other hand, removes redundant or insignificant parameters, further reducing the model size and computation load~\cite{han2016deepcompressioncompressingdeep}. Knowledge distillation involves training a small student model to mimic the behavior of a large teacher model, often achieving a student with a smaller footprint and comparable performance to those of the big teacher model~\cite{hinton2015distilling}. Additionally, efficient architecture designs introduce new layers types and reduce the number of parameters to optimize models performances~\cite{howard2017mobilenets, iandola2016squeezenet}.
Together, these methods significantly decrease memory usage and computation, while maintaining the model's accuracy within an acceptable range~\cite{8578384}.

Deploying these optimized models on MCUs also requires the development of optimized ML engines~\cite{alajlan2022tinyml}.
Two main types of engine address this problem: compiler-based and interpreter-based~\cite{dutta2021tinyml}.
The choice between the two depends on the specific application's requirements.
The interpreter-based approach provides flexibility and dynamic behavior but can be computationally expensive and less memory-efficient, while the compiler-based approach provides optimized and memory-efficient code but is less dynamic and more resource-intensive at compile-time. MicroFlow has been developed using a compiler-based approach in favour of runtime efficiency. Different solutions are discussed and compared in Sec.~\ref{sec:tinyml_frameworks}

\subsection{Programming Languages for TinyML Applications}

A programming language for embedded systems must provide low-level control of hardware resources, real-time responsiveness, and efficient memory management. Another crucial aspect is \emph{memory safety}, which refers to the protection of a program's memory from errors such as buffer overflows, use-after-free, and dangling pointers.
In recent years, memory safety has become a critical concern in software development, particularly for low-level programming languages.
The Microsoft Security Response Center~(MSRC) reported in 2019 that $70\%$ of all security vulnerabilities were caused by memory safety issues~\cite{msrc-2019-bluehat}.
Similarly, in 2020, a report from Google showed that $70\%$ of all severe security bugs in Google Chromium were caused by memory safety problems~\cite{chromium-2020-memory-safety}.
However, typical programming languages used for TinyML can be either inefficient (e.g. Python) or memory unsafe~(e.g. C/C++) requiring programmers to manually manage the memory usage.
With these languages, programmers must carefully follow memory safety best practices and use static analysis tools to detect potential issues.
These issues are particularly severe in bare-metal embedded devices, which lack the protection and abstractions provided by an operating system~(OS).

Memory-safe languages, instead, provide features such as automatic memory management, safe pointer arithmetic, and bounds checking, which significantly reduce the risk of memory errors.
However, memory-safe languages often rely on mechanisms such as garbage collection, which can introduce additional overhead and therefore impact performance~\cite{grgic2018comparison}.
It is therefore important to consider the trade-off between safety and performance when choosing a programming language for embedded systems.

Microflow has been developed using Rust, a general-purpose programming language introduced by Mozilla in 2010~\cite{klabnik2023rust}, which has gained popularity in recent years due to its unique features. It is increasingly being used in a variety of applications, including the Android Open Source Project~(AOSP)~\cite{rust-in-android-platform} and the Linux Kernel~\cite{lkml-linux-6.1}.
Rust is designed to provide the efficiency and control capabilities of low-level languages, like C and C++, while prioritizing memory and thread safety~\cite{jung2021safe}.
Instead of relying on garbage collection or manual memory management, Rust uses a system of ``ownership'' and ``borrowing'' to ensure that memory is allocated and deallocated safely and efficiently.
In this way, it is possible to write high-performance code without sacrificing safety or stability, making Rust a fast and memory-safe programming language~\cite{plauska2022performance}.
Moreover, it offers several other benefits for embedded systems~\cite{borgsmuller2021rust}.
In particular, the ownership mechanism ensures at compile-time that peripherals and I/O lines are correctly configured and used in a mutually exclusive way.
Compared to other programming languages like C/C++, Rust also excels in zero-cost abstraction and concurrency safety, leading to more expressive code while maintaining high runtime efficiency and robust compile-time guarantees~\cite{the-rust-programming-language}.
Furthermore, its rich ecosystem and active community also make it possible to develop portable libraries for TinyML easily.
Indeed there are many available and actively maintained libraries that offer non-standard support (i.e. targeting embedded platforms) and are useful for machine learning. The most popular example, mentioned later in the paper, is probably \texttt{nalgebra}\footnote{\url{https://nalgebra.org}}.
All this motivated the choice of Rust for MicroFlow's implementation, offering robustness and efficiency for bare-metal TinyML.

\subsection{Existing TinyML Frameworks} \label{sec:tinyml_frameworks}

There has been a growing interest in TinyML in recent years, with a number of software frameworks for research and practical applications.
Table~\ref{tab:inference-engines-summary} summarises the main features of some TinyML frameworks compared to MicroFlow.
Among these, TensorFlow Lite for Microcontrollers\footnote{\label{foot:tflm}\url{https://github.com/tensorflow/tflite-micro}}(TFLM) is a popular inference engine written in C++ and developed by Google~\cite{MLSYS2021_d2ddea18}.
It is built on top of TensorFlow Lite\footnote{\url{https://www.tensorflow.org/lite}} and is designed to be lightweight and efficient.
TFLM supports a wide variety of ML models and many of the commonly used operations, such as convolution, pooling, and fully connected layers.
Thanks to its popularity, the framework has inspired numerous projects that explore its potential further.
One such project is \emph{MicroNets}~\cite{MLSYS2021_c4d41d96}, which focuses on optimizing standard NNs to enable efficient inference using TFLM.
However, TFLM uses an interpreter-based approach, which is less efficient and requires a significant amount of memory.
Moreover, the framework supports only a limited number of 32-bit architectures.

\begin{table}
	\centering
	\def\arraystretch{1.2}
 \small
	\begin{tabular}{|l|c|c|c|l|l|c|}
		\hline
		\textbf{Framework} & \textbf{Open-source} & \textbf{Interpreter} & \textbf{Compiler} & \textbf{Language} & \textbf{min MCU} & \textbf{Bare-metal}\\ \hline
		TFLM\textsuperscript{\ref{foot:tflm}} & \textbullet & \textbullet & & C++ & 32-bit & \textbullet\\ \hline
		ELL\textsuperscript{\ref{foot:ell}} & \textbullet & & \textbullet & C++ & 32-bit & \textbullet\\ \hline
		ARM-NN\textsuperscript{\ref{foot:armnn}} & \textbullet & \textbullet & & C++ & 32-bit & \\ \hline
		Plumerai\textsuperscript{\ref{foot:plumerai}} & & & \textbullet & C++ & 32-bit & \textbullet\\ \hline
		uTensor\textsuperscript{\ref{foot:utensor}} & \textbullet & & \textbullet & C++ & 32-bit & \textbullet\\ \hline
		Tract\textsuperscript{\ref{foot:tract}} & \textbullet & \textbullet & & Rust & 32-bit & \\ \hline \hline
		\textbf{MicroFlow} & \textbullet & & \textbullet & Rust & 8-bit & \textbullet \\ \hline
	\end{tabular}
	\caption{Summary of the major features among different TinyML inference engines.}
	\label{tab:inference-engines-summary}
\end{table}

Embedded Learning Library\footnote{\label{foot:ell}\url{https://microsoft.github.io/ELL/}}(ELL) is a C++ library by Microsoft designed for ML models on resource-constrained devices.
Unlike TFLM, ELL adopts a compiler-based approach, which makes it more efficient and suitable for small embedded devices.
However, ELL is currently limited to a small number of ML algorithms and devices.
While this may be sufficient for some applications, it is not suitable for developers who require more advanced or specialized models.

ARM-NN\footnote{\label{foot:armnn}\url{https://www.arm.com/products/silicon-ip-cpu/ethos/arm-nn}} is an open-source C++ software library designed for accelerating ML development on ARM-based devices.
One of the key benefits of ARM-NN is its ability to provide optimized code for ARM MCUs, which are widely used in embedded systems.
The library is designed to work with a variety of ML frameworks, including TensorFlow, Caffe, and PyTorch, making it a flexible option (although it does not support all the features or functionalities of those frameworks).
Unfortunately, one obvious limitation is that it is only suitable for ARM-based devices.

Plumerai\footnote{\label{foot:plumerai}\url{https://plumerai.com/}} is a startup company that specializes in developing ML tools and platforms for embedded systems.
One of Plumerai's key products is a inference engine that combines state-of-the-art techniques, such as Binarized Neural Networks~(BNNs)~\cite{courbariaux2016binarized}, with a compiler-based approach to obtain a fast and efficient solution.
However, the Plumerai software is written in C++ and it is currently proprietary (i.e. closed-source), which limits the ability to understand how the software works, make modifications or improvements, and address potential security vulnerabilities.

uTensor\footnote{\label{foot:utensor}\url{https://utensor.github.io/website/}} is an extremely light-weight ML inference framework built on TensorFlow and optimized for ARM targets.
The framework is implemented in C++ and leverages the ARM Compute Library for optimized matrix operations, suitable only for ARM-based microcontrollers.
However, it does not offer support for complex ML models or for those requiring more advanced optimization techniques.

Finally, Tract\footnote{\label{foot:tract}\url{https://github.com/sonos/tract}} is an inference engine written in Rust and developed by Sonos.
Unlike other Rust-based solutions that contain bindings to C/C++ libraries -- effectively voiding the memory-safe guarantee of the language -- Tract is completely self-contained and independent from non-Rust components.
However, one of its main limitations is that it requires the Rust Standard Library, which is not available for bare-metal systems without an OS.
This limitation affects the portability of Tract, making it suitable only for devices with significant processing power and memory resources.

\section{System Design and Components}
\label{sec:design}

Strengths and limitations of current TinyML systems in the related work helped to identify a set of core design principles for MicroFlow, which are discussed next. These are followed by a presentation of its general structure and main software components, namely the Compiler and the Runtime modules.

\subsection{Design Goals}

Based on the review of existing frameworks, a number choices and principles have guided the design of MicroFlow, mainly to address portability, efficiency, robustness, and scalability of the proposed solution.

\paragraph{Portability}
The embedded ecosystem is very fragmented, with many vendor-specific frameworks and architectures.
It is therefore very difficult to develop a single software package that works efficiently and seamlessly on all the available devices.
When using traditional programming languages, such as C or C++, the challenge is further accentuated because these languages are defined by common standards but implemented by different compilers, each with its own features and peculiarities.
The Rust programming language, instead, provides a more convenient way to build portable software, since its ecosystem is more centralized.
Moreover, the language is defined by the compiler itself, providing a single, official instance that takes care of building the code for the target architecture.
Rust also comes with official built-in toolchain manager\footnote{\url{https://github.com/rust-lang/rustup}} and package manager\footnote{\url{https://github.com/rust-lang/cargo}}, enabling the creation of portable software without worrying about vendor-specific details.

\paragraph{Efficiency}
One of the main challenges of embedded systems programming is the limited amount of available resources.
To this end, MicroFlow adopts a compiler-based approach that benefits from advanced optimization techniques and static analysis, resulting in improved performance and memory efficiency.
Additionally, Rust features such as low-level control, zero-cost abstractions, and minimal runtime overhead, contribute to the overall efficiency of the proposed framework.

\paragraph{Robustness}
NN-based inference can be very memory-intensive and requires heavy matrix manipulation and processing.
This can increase the likelihood and negative effects of memory-related bugs, such as index-out-of-bounds, segmentation faults, stack overflows, and so on.
The strong typing system offered by Rust ensure, at compile-time, that all the operations made during execution are memory-safe.
In addition, it is also possible to use external libraries\footnote{In Rust, each unit of software is shipped inside a so-called \emph{crate}. All the crates are collected and available in the central Rust Crate Registry (\url{https://crates.io}).} (i.e. ``crates'') that are fully written in Rust, so the whole program follows the strict language's rules of ownership and borrowing, ensuring its safe execution.

\paragraph{Scalability}
NNs are are essentially computational graphs consisting of sequences of operators.
The sequence, number, and hierarchy of the operators define the architecture of the model.
As ML models evolve and become more complex, there is a constant need to introduce new operators or custom operations with enhanced capabilities.
MicroFlow inference engine provides the necessary infrastructure to easily implement, integrate, and scale up these operations by implementing modular and reusable code, as detailed in Sec.~\ref{sec:operators}.

\subsection{Software Structure}

\begin{figure}
    \centering
    \includegraphics[width=.7\textwidth]{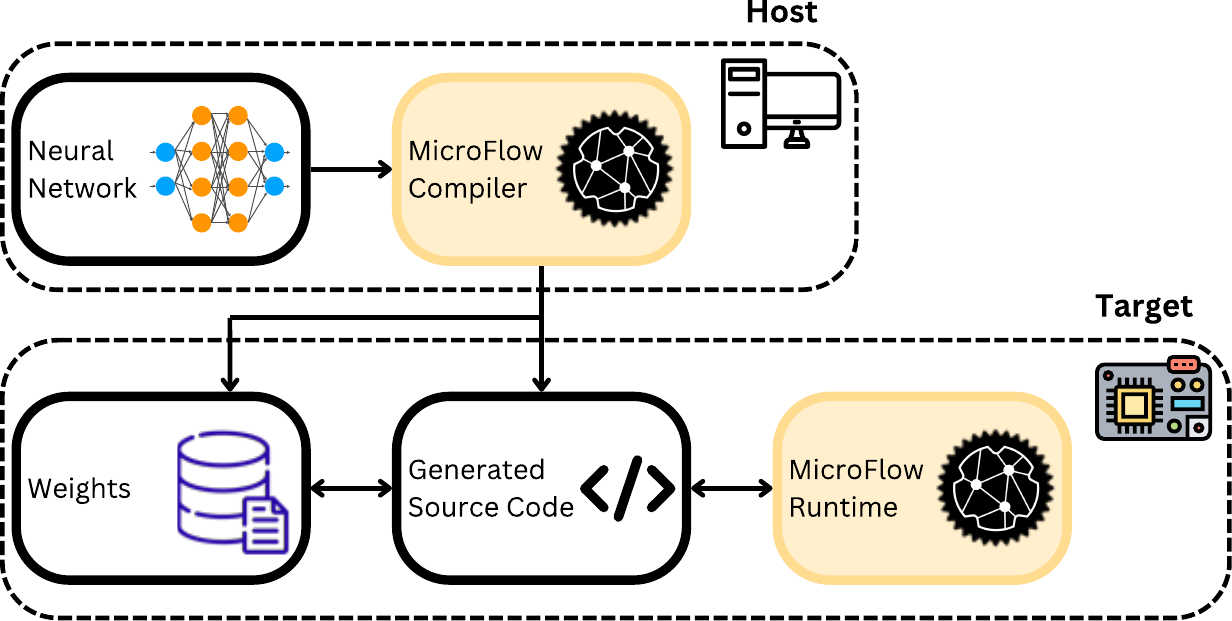}
    \caption{Overview of the MicroFlow framework. Given a Neural Network, the host machine generates the Source Code and the network encoded Weights using the MicroFlow Compiler. The target embedded system executes the model using the MicroFlow Runtime module.}
    \label{fig:structure-overview}
\end{figure}

The high-level structure of the proposed framework is shown in Fig.~\ref{fig:structure-overview} and includes two major components: the MicroFlow Compiler, which resides on the host machine, and the MicroFlow Runtime, which resides on the target MCU.
The goal is to delegate as much work as possible to the compiler, creating a lightweight runtime process that performs only the essential computations during program execution.
The compiler is also responsible for analysing the model to determine the minimum amount of memory that must be statically allocated for runtime inference.

Every operator in MicroFlow derives from a template consisting of two parts: the {\em parser} and the {\em kernel}.
The parser runs statically in the Compiler, preprocessing the model and preparing the weights for the runtime execution.
The kernel runs on the Runtime component and takes care of the actual computation of the operator, propagating the input to the output.
Each operator is isolated from the others, communicating only through input and output interfaces, and leaving no memory trace after its execution.

Although MicroFlow is designed to support a multitude of NN operators, only the most commonly used are currently supported:
FullyConnected,
Conv2D,
DepthwiseConv2D,
AveragePool2D,
Reshape,
ReLU,
ReLU6,
and Softmax.
With these operators, it is possible to implemenet a wide range of NN models, including Feedforward Neural Networks~(FNNs) and Convolutional Neural Networks~(CNNs). These are further discussed in Sec.~\ref{sec:operators} and tested in Sec.~\ref{sec:evaluation}.
The scalable structure of MicroFlow enables future improvements and implementations of new operators.

\subsection{MicroFlow Compiler}

As discussed in the related work, typically there are two main methods used to develop inference engines for embedded systems:
\begin{itemize}
\item In the interpreter-based approach, the inference engine dynamically parses and convert the model into machine instructions at runtime.
This approach is generally more flexible and requires shorter compilation times, but it has some drawbacks.
First of all, interpreting at runtime can introduce a significant performance overhead due to many additional operations, such as model parsing, type-checking, and memory allocation.
The latter in particular introduces several risks such as memory leaks, heap fragmentation, and security vulnerabilities.
Moreover, the interpreter itself can take up a significant amount of memory, regardless of the size of the network, and it cannot be optimized since the network size is not known a-priori.

\item In the compiler-based approach, the model is translated into machine code that can be executed directly by the processor.
Model's parsing and evaluation are done at compile-time (on a host system), which can be relatively time-consuming and resource-intensive. Obviously, any changes to the model require recompilation, but the advantage is to have optimized code that can be executed quickly and efficiently.
Memory management is also handled during the compilation stage, with all the memory allocations done statically.
This avoids any risks related to dynamic memory management and reduces the inference engine's footprint, even because the latter is proportional to the original model's size, so small models can run on highly constrained devices.
Moreover, the parts of the model that are not required at runtime (such as operator identifiers, names, and version numbers) can be stripped away, resulting in an even smaller binary size.
\end{itemize}

\begin{figure}
    \centering
    \includegraphics[width=.7\textwidth]{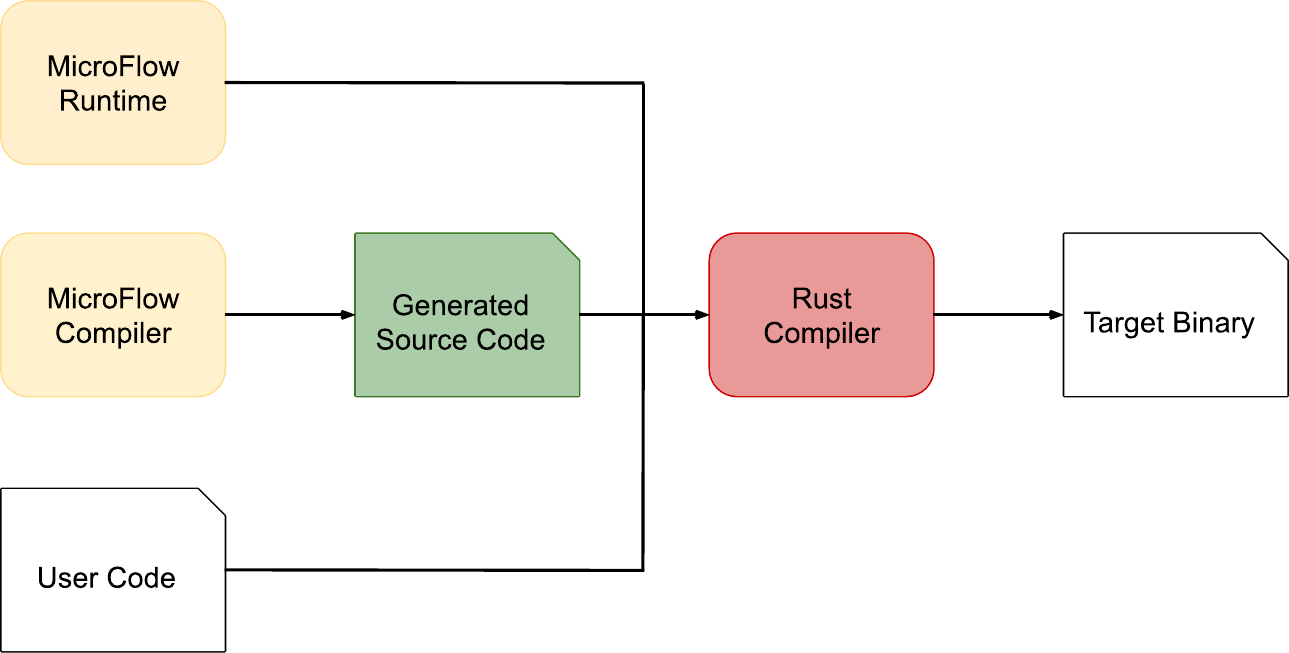}
    \caption{MicroFlow's compilation steps. The MicroFlow Compiler generates the Source Code that is eventually built by the Rust Compiler, together with the Microflow Runtime and the User Code, to produce the Target Binary.}
    \label{fig:compilation-flow}
\end{figure}

The latter approach best suites MicroFlow's design goals. In particular, MicroFlow Compiler processes a given NN model and generates the necessary inference code.
The implementation is structured as a sub-crate of the main \texttt{microflow} crate, called \texttt{microflow-macros}, due to its extensive use of Rust macros.
The compilation steps are illustrated in Fig.~\ref{fig:compilation-flow}.
The MicroFlow Compiler performs the first stage of the building process, producing the source code that is finally built by the Rust Compiler.
The two primary components of the MicroFlow Compiler are the set of Rust macros used to generate the source code and the parsing process that analyzes the model. Both are discussed in the next subsections.

\subsubsection{Macros}

Rust macros play a vital role in MicroFlow.
They provide a powerful mechanism for code generation and \emph{metaprogramming}, enabling developers to write programs that generate or manipulates code at compile time.
In Rust, there are two types of macros: declarative macros and procedural macros.
Declarative macros, also known as \emph{macro\_rules}, allow for pattern matching and substitution within the code.
They are defined using the \texttt{macro\_rules!} keyword and are expanded at compile-time.
Procedural macros, on the other hand, enable code generation and transformation by implementing custom logic.
They are typically defined as separate crates (in MicroFlow, they are defined in the \texttt{microflow-macros} crate) and are invoked using attributes or function-like syntax.
Procedural macros are expanded at the early stage of compilation, as shown in Fig.~\ref{fig:compilation-flow}, and are more powerful than declarative macros.

\begin{figure}[b]
    \centering
    \includegraphics[width=\textwidth]{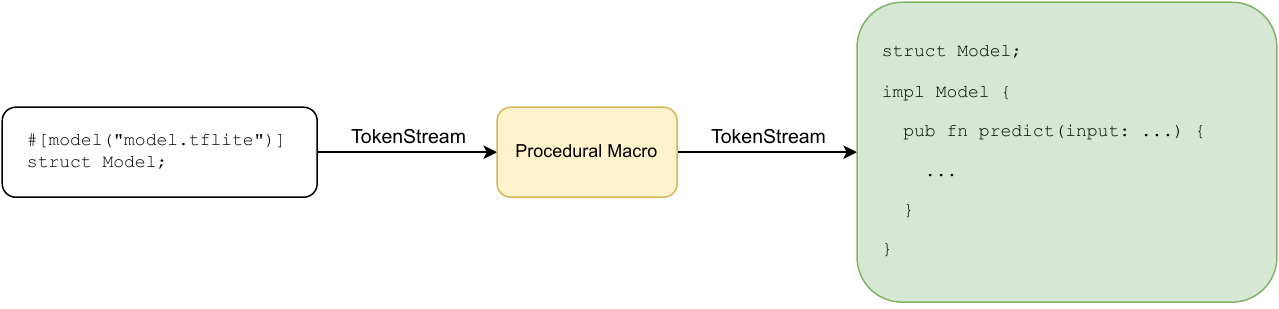}
    \caption{Expansion of the macro. The input tokens are expanded by the procedural macro according to the model.}
    \label{fig:macro-expansion}
\end{figure}

A macro receives as input a stream of tokens representing the macro invocation (along with its parameters) and outputs a stream of tokens representing the generated code.
In this case, an attribute-like procedural macro~\cite{TheRustReference2024}, named \texttt{model}, was implemented to annotate a \texttt{struct} and bind it to a NN model.
The macro takes as input the path to the model in the TensorFlow Lite~(TFLite) format and generates a \texttt{predict()} function that, when called, performs inference on the given model.
The parsing of the model and the generation of the source code for the function are entirely computed during compilation.
The \texttt{predict()} function is embedded in the source file by the macro expansion, and it is subject to all the operations of the compiler, including memory safety checks and optimizations.
An example of the code expansion can be seen in Fig.~\ref{fig:macro-expansion}.
Overall, macros are the core component of the compiler, and they act as the entrypoint for the entire software.
They start the parser and manage the generation of the runtime calls.

\subsubsection{Parsing} \label{sec:parsing}

The parsing phase analyses the input model and generates the output code plus the memory structures, accordingly.
The input of the parser comes from the macro, and it is the path of the model relative to the root of the crate.
Although MicroFlow accepts, as input, NN models in the TFLite format,
other formats (e.g. ONNX) could be included simply by expanding the parser.
Under the hood, TFLite uses the \emph{FlatBuffers} serialization format, which provides a lightweight and efficient solution for serializing and deserializing structured data suitable for embedded applications~\cite{MLSYS2021_d2ddea18}.
For this, FlatBuffers relies on a schema definition of the data's structure and layout.
Therefore, there is not a single parser for FlatBuffers, but it depends on the schema.
Fortunately, the format includes also a powerful code generation tool called \texttt{flatc}, which takes a FlatBuffers schema as input and automatically generates a parser handling serialization and deserialization of FlatBuffers files, based on the given schema.

MicroFlow's parser starts by invoking the FlatBuffers deserializer generated by \texttt{flatc}, then proceeds extracting the operators of the NN, along with all the tensor dimensions, content, and relations.
Subsequently, it generates an internal representation of the model by constructing a series of operators, each one associated with its respective parameters, such as the input/output tensors, weights, activation function, and other relevant attributes.
Each operator contains also the stream of tokens needed by the macro to generate its runtime call.
For example, the FullyConnected operator in the internal representation will contain the tokens that, once included in the generated source code, call the \texttt{fully\_connected()} function at runtime, with all the required arguments.

\begin{figure}
    \centering
    \includegraphics[width=0.85\textwidth]{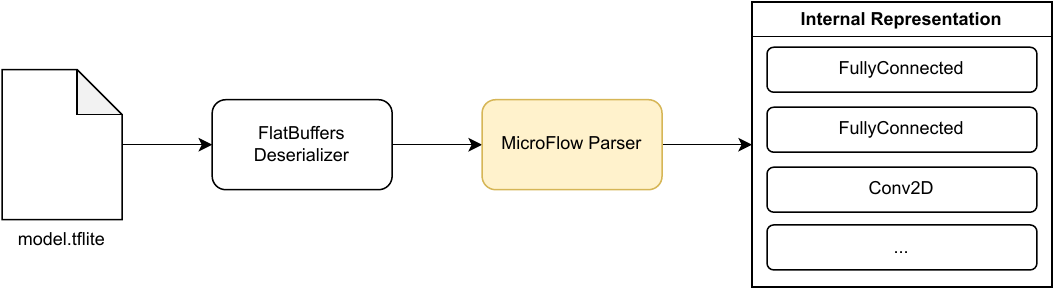}
    \caption{Parsing example. The input file is deserialized and parsed to build the internal representation.}
    \label{fig:parsing-execution}
\end{figure}

An example of the parsing sequence is illustrated in Fig.~\ref{fig:parsing-execution}.
The internal representation captures the structure and characteristics of the model, enabling further processing and manipulation for efficient execution of the NN inference.
This internal representation is lossless, that is, it contains a reversible representation of the original NN. Therefore, the performance and accuracy of the parsed model are equal to those of the input one.
Once this internal representation is built, along with the sequence of operators involved, the parser proceeds with the pre-processing phase.

\subsubsection{Pre-processing}

The pre-processing phase of the MicroFlow Compiler plays a crucial role to reduce the load at runtime by performing calculations and optimizations on constant values during compile-time.
During this phase, the \texttt{preprocess()} function that is present in each operator of the internal representation is invoked.
By offloading constant calculations to the pre-processing phase, the runtime performance is improved as it avoids redundant operations and reduces the computational overhead during inference.

The pre-processable part of each operator is obtained by analyzing its mathematical properties.
In fact, everything in the operator's formula that does not depend on its input can be considered in this phase.
Sec.~\ref{sec:operators} provides a detailed explanation of the extraction of constant values from the operator's transfer functions.
Finally, the values computed during this pre-processing phase are stored in custom tensors. The latter are passed as arguments to the operator's kernel in the MicroFlow Runtime component, ensuring that the pre-computed values are readily available for efficient execution of the inference process.

\subsection{MicroFlow Runtime}

The MicroFlow Runtime is the second main component of the software.
As the name suggests, it contains all the implementation of the operators and, more generally, everything that is executed on the target MCU.
A key difference between Compiler and Runtime components is that the latter cannot rely on the Rust standard library.
In fact, on bare-metal MCU, there is no OS and the software can only access the most essential structures and components of the language from the core library.
Since the latter is a subset of the standard library, the MicroFlow Runtime can run both on bare-metal and OS-based platforms.

This component must be operated efficiently in terms of computational performance and memory management.
The runtime functions are called by the code generated by the compiler, after some static checks are performed to ensure reliable execution.
In particular, the MicroFlow Runtime receives the sequence of operators to execute along with the tensors.
The model does not need to be evaluated at runtime, unlike other solutions in the literature.
For example, the TFLM inference engine consists only of a runtime module, which is an interpreter~\cite{MLSYS2021_d2ddea18}.
This causes overhead because all the operations carried out by the compiler need to be carried out at runtime by the interpreter instead.

Another responsibility of the MicroFlow Runtime is to manage memory allocation.
However, since the model is fully analyzed before execution, the memory needed to perform inference is statically defined.
As a result, the runtime module knows in advance the exact amount of memory needed and the specific locations where tensors should be stored.
This enables a precise allocation of memory resources, optimizing their use and minimizing overhead.

\subsubsection{External Libraries}

The Runtime module relies on external libraries to perform matrix operations and manipulations.
These libraries must be independent from the standard one and completely written in Rust to ensure memory safety.
Also, MircoFlow needs a library that can be used with static memory allocation and generic types, or {\em generics} (discussed next).
To this end, the \texttt{nalgebra}\footnotemark[2] crate is a powerful linear algebra library, fully written in Rust -- and thus memory-safe -- that provides a comprehensive set of tools and structures for mathematical operations involving vectors, matrices, and other geometric entities.
In line with MicroFlow's goals, it is designed to be efficient, generic, and easy to use.
One of the key features of \texttt{nalgebra} is the support for both fixed-size and dynamically-sized matrices and vectors.
Moreover, it puts a strong emphasis on generics, which play a significant role in the proposed software since they enhance the versatility and adaptability of the Runtime part, making it suitable for a wide range of models.

\subsubsection{Generics}

Generic programming is a fundamental concept in Rust that allows the creation of highly versatile and reusable code.
It enables the definition of functions, structs, and traits that can work with multiple data types, providing a high level of flexibility and abstraction.
By leveraging generics, the MicroFlow Runtime component can provide a general interface for working with various NN models, allowing users to easily integrate different architectures and quantization types without sacrificing performance or safety.
Rust offers also {\em const generics}, i.e. generic arguments that range over constant values, allowing types or functions to be parameterized by integers.
Like the other generics, the const ones are also evaluated and expanded at compile-time.

The MicroFlow Runtime is built on top of generic types and const generics.
The first makes it possible to provide a single definition of a runtime function that works for different types.
Const generics instead allow to parameterize a function over some numerical constants obtained at compile time. In this case, the numerical constants are the tensor shapes. By doing so, there is no need to provide different function definitions for different tensor sizes, and no runtime execution has to be wasted accessing the tensor struct to get its size, or accessing a function parameter with the tensor size. Every numerical constant is pre-filled by the Rust compiler.
Moreover, the types and the const generics can be restricted and correlated with each other to ensure correctness at compile-time.

\section{Memory Management}\label{sec:memory_management}

Memory management is a critical aspect to perform inference on resource-constrained MCUs.
Efficient memory utilization is also essential to ensure optimal performance, minimize memory footprint, and avoid issues such as memory leaks and crashes.
In this section, the techniques and strategies adopted in MicroFlow are explained.

\subsection{Ownership}

To understand how memory is managed in MicroFlow, it is useful to first understand Rust's memory management, particularly its {\em ownership} concept~\cite{klabnik2023rust}. In Rust, every value has a single owner responsible for its memory. Ownership can be transferred when a value is assigned or passed to a function, ensuring a clear, conflict-free memory management system.
Rust then enforces {\em borrowing} rules to prevent data races by allowing either multiple immutable references or a single mutable reference to a value, thus avoiding concurrent modifications. Memory is automatically deallocated when an owner goes out of scope, preventing memory leaks and eliminating the need for manual memory management.
Rust allows borrowing through references, which can be immutable or mutable, enabling safe data sharing without transferring ownership. The Rust compiler enforces these ownership and borrowing rules at compile-time, raising errors if violations occur, ensuring memory safety and preventing unsafe operations.

In MicroFlow, these concepts are used to ensure memory efficiency.
First of all, since all the tensor dimensions are known at compile-time, the entire execution at runtime does not require any dynamic allocation on the heap.
This results in an optimal memory utilization since everything is allocated on the stack and freed after use.
By doing so, problems such as memory fragmentation and dangling pointers are avoided.
Moreover, the code becomes more portable and user-friendly since the programmer does not have to provide a global heap allocator or a memory arena.
Instead, the needed memory is statically defined and allocated on the stack.

\begin{figure}
    \centering
    \includegraphics[width=0.7\textwidth]{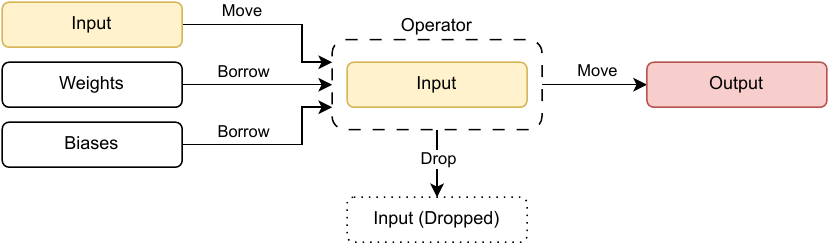}
    \caption{Example of ownership propagation during the execution of an operator. The input tensor is transferred to the operator, which receives ownership and releases the tensor after execution.}
    \label{fig:ownership-propagation}
\end{figure}

MicroFlow handles the transfer of responsibilities according to the following mechanism.
Each operator takes ownership of the input tensor and immutably borrows the others.
The operator then moves the output tensor to the input of the next operator, which in turn takes ownership of the tensor.
This mechanism ensures that the lifetime of the input tensor is bound to the operator, dropping the tensor once all the values have been propagated to the output.
Therefore, at any point in time, the current working operator uses the minimum amount of memory.
For the other tensors, such as weights and biases, ownership transfer is not necessary because they have constant values that are never dropped.
Instead, since they are only read, they can be efficiently accessed by a borrow request (i.e. an immutable reference).
An example diagram of this mechanism is illustrated in Fig.~\ref{fig:ownership-propagation}.

\subsection{Static Allocation}

Everything in MicroFlow is allocated on the stack.
By doing so, the memory used by operators will peak when the most memory-intensive operators are executed, then it will be automatically freed.
Therefore, once the entire inference process is finished,
the memory allocated by the engine is null.

An interpreter-based system like TFLM, instead, allocates the tensor arena for the entire execution of the inference process.
The arena's size is constant and it is not freed after use. It must also be big enough to accommodate the most memory-intensive operator.
Moreover, in TFLM the programmer is responsible for manually allocating and deallocating memory, as well as determining the appropriate amount to be used.
This can result in suboptimal memory allocations and potential runtime errors (i.e. in case the user allocates either too little or too much memory).

\subsection{Paging}

With the ownership mechanism described in the previous section, an entire NN layer can be loaded in RAM during execution.
However, certain MCU have limited RAM, which can be a challenge for large layers.
For example, the ATmega328 of the popular Arduino Uno boards has 32kB of Flash memory and only 2kB of RAM~\cite{ATmega328}.
This is not enough to perform inference with a NN's dense layer of 32 fully connected neurons, since the memory required would be approximately\footnote{The calculation considers weights ($32\times32$), 32-bit signed integer accumulators ($4\times32\times32$), and vectors containing biases, input, and output ($3\times32$).} 5kB.
Although the Flash memory can hold the entire NN, its size would cause a stack overflow.

\begin{figure}
    \centering
    \includegraphics[width=.45\textwidth]{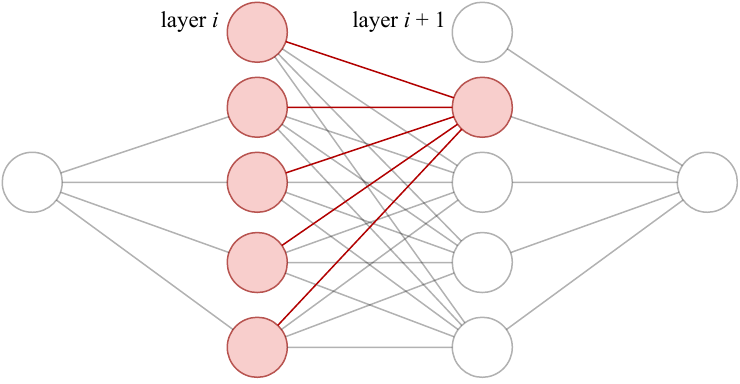}
    \caption{Example of MicroFlow paging for a fully connected layer. The page contains the information of the elements highlighted in red, including 4 inputs, 4 weights, 1 bias, and 1 output.}
    \label{fig:paging-example}
\end{figure}

A few solutions exist to reduce the need of storing intermediate results~\cite{TFLiteMemOpt,EdgeImpulseMemOpt}. In MicroFlow, such limitation is overcome by dividing the layer into \emph{pages} and loading them in RAM one at a time.
This approach allows for efficient memory utilization and ensures that the MCU's RAM is not overwhelmed.
A layer's page includes all the information related to the connections from the units of layer $i$ to one unit of layer $i+1$, as illustrated in Fig.~\ref{fig:paging-example}.
For the previous example, dividing the layer into 32~pages results in a manageable RAM usage of 163~bytes.
Dividing and loading the layer in pages, however, increases the execution time.
Therefore, in situations where memory resources are limited and slow inference times are acceptable, the paging approach can be a viable solution.
On the other hand, if memory constraints are less stringent and faster inference times are crucial, loading the entire layer at once remains the preferred option.

\subsection{Stack Overflow Protection}

Although Rust is known for its emphasis on memory safety, this is not guaranteed for bare-metal programs in case of stack overflow.
For example, on the broadly used ARM Cortex-M architectures,
the stack can grow too large and collide with the region containing all the program's static variables, overwriting them and causing undefined behaviors.

A solution is to use a ``flipped'' memory layout
by placing the stack {\em below} the aforementioned region, thus possibly colliding only with the boundary of the allocated RAM space.
On ARM Cortex-M, trying to read or write past this boundary produces a hardware exception, which can be handled by Rust.
To this end, MicroFlow utilizes the \texttt{flip-link}\footnote{\url{https://crates.io/crates/flip-link}} crate, a replacement of the default Rust linker that can flip the memory layout.
Currently, this crate works only for Cortex-M architectures, which are therefore the only ones with stack overflow protection in MicroFlow, but more platforms will be supported in the future.

\section{MicroFlow Operators}\label{sec:operators}

NN processes are often represented by a sequence of operations in a computational graph.
As reported in Table~\ref{tab:operators} and documented on the official Rust documentation platform\footnote{\label{foot:docs}\url{https://docs.rs/microflow/latest/microflow/}}, MicroFlow provides the most common ones to implement FNNs and CNNs, including FullyConnected, Conv2D, DepthwiseConv2D, AveragePool2D, Reshape, ReLU, ReLU6, and Softmax operators.
Apart for Reshape, which simply re-arranges the shape of a tensor, all the operators need to be quantized before using them for TinyML applications: that is, floating point numbers have to be converted into integer ones, which should be still accurate enough to perform inference correctly.
After quantization, each number is mapped according to the following equation:
\begin{equation}
\label{eq:dequantization}
    r = S(q - Z)
\end{equation}
where $r$ is the original floating-point value, $q$ is the quantized fixed-point value, and $S$ and $Z$ are the quantization parameters, namely \emph{scale} and \emph{zero point}, respectively.
The quantization takes place before deployment, either during or after training.
Its parameters are calculated based on a representative sample of the input data.
During inference, the engine uses the quantized data together with the quantization parameters.
Some of the most popular ML frameworks provide support for quantization~\cite{liang2021pruning}.
Among these, TensorFlow Lite is perhaps the most widely used, since it provides a set of tools and libraries for developers to deploy and run ML models on mobile and embedded devices.
It allows to train, quantize, and save a model for embedded applications in a dedicated TFLite format, which is the one adopted by MicroFlow -- although any other quantization tool compatible with this format could be used.

\begin{table}
	\centering
	\def\arraystretch{1.2}
        \small
	\begin{tabular}{|l|l|l|}
    \hline
		\textbf{Operator} & \textbf{Quantization} & \textbf{Tensor Type} \\ \hline
		\emph{FullyConnected} & Eq.~(\ref{eq:fullyconnected}) & \emph{Tensor2D} \\ \hline
        \emph{Conv2D} & Eq.~(\ref{eq:conv2d}) & \emph{Tensor4D} \\ \hline
        \emph{DepthwiseConv2D} & Eq.~(\ref{eq:depthwiseconv2d}) & \emph{Tensor4D} \\ \hline
        \emph{AveragePool2D} & Eq.~(\ref{eq:averagepool2d}) & \emph{Tensor4D} \\ \hline
        \emph{Reshape} & -- & \emph{Tensor2D, Tensor4D} \\ \hline
	\end{tabular}
        \hfil
	\begin{tabular}{|l|l|}
		\hline
        \textbf{Activation Function} & \textbf{Quantization} \\ \hline
        \emph{ReLU} & Eq.~(\ref{eq:relu}) \\ \hline
        \emph{ReLU6} & Eq.~(\ref{eq:relu6}) \\ \hline
        \emph{Softmax} & Eq.~(\ref{eq:softmax}) \\ \hline
	\end{tabular}
	\caption{Operators and activation functions implemented currently supported by MicroFlow (v0.1.3) and documented on the Rust documentation platform\textsuperscript{\ref{foot:docs}}. The tables include references to the quantized equations discussed in Sec.~\ref{sec:operators}. Note that the {\em Reshape} operator does not require quantization.}
	\label{tab:operators}
\end{table}

\begin{figure}
    \centering
    \includegraphics[width=.4\textwidth]{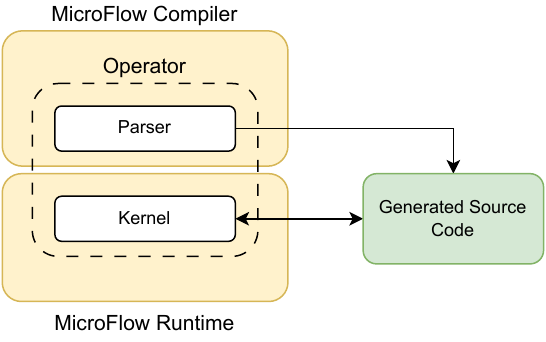}
    \caption{Operator's components. The parser resides in the MicroFlow Compiler and contributes to the generated code. The kernel resides in the MicroFlow Runtime and contributes to the inference.}
    \label{fig:operator-components}
\end{figure}

Once an operator has been quantized, its implementation in MicroFlow follows the design described in Sec.~\ref{sec:design}.
The operator is split into two components: the parser, which runs on the compiler, and the kernel, which runs at runtime.
The goal of the parser is to facilitate the kernel's job by preparing the input, output, and intermediate tensors, and by pre-processing the constant values.
The goal of the kernel is to propagate the input to the output in the most efficient way.
An overview of the relation between the operator's components is shown in Fig.~\ref{fig:operator-components}.

Unfortunately, to the best of our knowledge, the mathematical derivation of the quantized operators used in MicroFlow is not available in the literature. This is essential, however, to distinguish which parts of the code can be generated at compile-time and which at run-time.
Therefore, the remaining part of this section presents the quantized formulae of the implemented operators, derived from the original notions of NN quantization~\cite{8578384}, and explains which constant terms are pre-computed by the MicroFlow Compiler.
Further details are also available in Appendix~\ref{appendix:operators}.

\subsection{FullyConnected}

The FullyConnected operator, also known as the dense or linear operator, is a key building block of many neural networks.
In this operator, each input element is multiplied by a corresponding weight and summed to other weighted inputs and biases.
The resulting sum is then passed through a non-linear activation function to produce the final output value.
More formally, given $X \in \mathbb{R}^{m \times n}$, $W \in \mathbb{R}^{n \times p}$, and $b \in \mathbb{R}^{p}$, representing respectively the input, weights, and biases of the operator, the output $Y \in \mathbb{R}^{m \times p}$ can be written as follows:
\begin{equation}
    Y_{i,j} = b_j + \sum_{k=1}^{n} X_{i,k} W_{k,j}
\end{equation}
By applying Eq.~(\ref{eq:dequantization}), the quantized version can be derived (details in Appendix~\ref{appendix:fullyconnected}):
\begin{equation}
\label{eq:fullyconnected}
    Y_{q,i,j} = z_Y + \frac{s_b}{s_Y} (b_{q,j} - z_b) + \frac{s_X s_W}{s_Y} \Bigg[ \bigg( \sum_{k=1}^{n} X_{q,i,k} W_{q,k,j} \bigg) - \bigg( z_W \sum_{k=1}^{n} X_{q,i,k} \bigg) - \bigg( z_X \sum_{k=1}^{n} W_{q,k,j} \bigg) + n z_X z_W \Bigg]
\end{equation}
where $X_q$, $W_q$, $b_q$, and $Y_q$ are the quantized versions of $X$, $W$, $b$, and $Y$, respectively, $s_X$, $s_W$, $s_b$, and $s_Y$ are the scales, while $z_X$, $z_W$, $z_b$, and $z_Y$ are the zero points.
During inference, the following terms are constant:
\begin{equation}
    z_Y + \frac{s_b}{s_Y} (b_{q,j} - z_b)  \qquad
    \frac{s_X s_W}{s_Y}  \qquad
    z_X \sum_{k=1}^{n} W_{q,k,j}  \qquad
    n z_X z_W
\end{equation}
therefore they can be pre-computed offline by the operator parsing phase of the compiler, reducing runtime overhead, while the remaining calculation to obtain $Y_q$ will be implemented in the operator's kernel.

\subsection{Conv2D}

The Conv2D (short for ``Convolutional 2D'') operator is a fundamental building block of CNNs.
It performs a convolution operation on an input tensor using a set of learnable filters.
Conv2D operators are commonly used for tasks such as image recognition, object detection, and image segmentation.
They capture local patterns and spatial relationships in the input data, allowing the NN to learn hierarchical representations and extract meaningful features.
Here tensors are composed by a set of matrices (\emph{batches}) containing multiple values (\emph{channels}), one for each position.
The convolutional filters are represented by the batches, while the channels are merged together by a dot-product during convolution.
The output tensor contains only one batch for each channel with the result of the convolution, applied at that position in the input matrix.
The quantization process is carried out as follows:
given $X \in \mathbb{R}^{m \times n \times c}$, $F \in \mathbb{R}^{m \times n \times c}$, and $b \in \mathbb{R}$, representing respectively an input region, a filter, and the bias, the output value $y \in \mathbb{R}$ for a given channel can be written as follows:
\begin{equation}
    y = b + \sum_{i=1}^{m} \sum_{j=1}^{n} \sum_{k=1}^{c} X_{i,j,k}  F_{i,j,k}
\end{equation}
After quantization, the following expression is obtained (details in Appendix~\ref{appendix:conv2d}):
\begin{equation}
\begin{split}
\label{eq:conv2d}
    y_q &= z_y + \frac{s_b}{s_Y} (b_q - z_b) + \frac{s_X s_F}{s_y} \Bigg[ \bigg( \sum_{i=1}^{m} \sum_{j=1}^{n} \sum_{k=1}^{c} X_{q,i,j,k} F_{q,i,j,k} \bigg) - \bigg( z_F \sum_{i=1}^{m} \sum_{j=1}^{n} \sum_{k=1}^{c} X_{q,i,j,k} \bigg) \\
    & \qquad - \bigg( z_X \sum_{i=1}^{m} \sum_{j=1}^{n} \sum_{k=1}^{c} F_{q,i,j,k} \bigg) + m~n~c~z_X z_F \Bigg]
\end{split}
\end{equation}
where $X_q$, $F_q$, $b_q$, and $y_q$ are the quantized versions of $X$, $F$, $b$, and $y$, respectively, $s_X$, $s_F$, $s_b$, and $s_y$ are the scales for $X$, $F$, $b$, and $y$, respectively, and $z_X$, $z_F$, $z_b$, and $z_y$ are the zero points for $X$, $F$, $b$, and $y$, respectively.
Since the following terms are constants, they can be pre-processed by the compiler:
\begin{equation}
    z_y + \frac{s_b}{s_Y} (b_q - z_b) \qquad
    \frac{s_X s_F}{s_y} \qquad
    z_X \sum_{i=1}^{m} \sum_{j=1}^{n} \sum_{k=1}^{c} F_{q,i,j,k} \qquad
    m~n~c~z_X z_F
\end{equation}

Note that the implementation of the Conv2D operator requires also an additional view extraction routine in the kernel to select the appropriate input elements used in each convolution, taking into account padding and stride (details in Appendix~\ref{appendix:conv2d}).

\subsection{DepthwiseConv2D}

The DepthwiseConv2D operator is a specific type of Conv2D operator, commonly used in efficient CNNs like MobileNet~\cite{howard2017mobilenets}, which
applies a separate filter for each input channel.
This means the operator performs depthwise convolutions, as its name suggests, where each channel is convolved independently.
The DepthwiseConv2D operator shares many properties and data structures with the conventional Conv2D.
However, the process is different:
instead of merging the channels with a dot-product, they are kept separate and convolved individually with the corresponding channels of the filter.
In particular, DepthwiseConv2D operates on a single batch with a 3D weight matrix, where the third dimension represents the weights associated to each channel.
Therefore, given $X \in \mathbb{R}^{m \times n}$, $W \in \mathbb{R}^{m \times n}$, and $b \in \mathbb{R}$, which represent the input region, the weights matrix for a given channel, and the bias, respectively, the output value $y \in \mathbb{R}$ can be written as follows:
\begin{equation}
    y = b + \sum_{i=1}^{m} \sum_{j=1}^{n} X_{i,j}  W_{i,j}
\end{equation}
By applying the usual quantization, the resulting output is the following one (details in Appendix~\ref{appendix:depthwiseconv2d}):
\begin{equation}
\begin{split}
\label{eq:depthwiseconv2d}
    y_q &= z_y + \frac{s_b}{s_y} (b_q - z_b) + \frac{s_X s_W}{s_y} \Bigg[ \bigg( \sum_{i=1}^{m} \sum_{j=1}^{n} X_{q,i,j} W_{q,i,j} \bigg) - \bigg( z_W \sum_{i=1}^{m} \sum_{j=1}^{n} X_{q,i,j} \bigg) \\
    & \qquad - \bigg( z_X \sum_{i=1}^{m} \sum_{j=1}^{n} W_{q,i,j} \bigg) + m~n~z_X z_W \Bigg]
\end{split}
\end{equation}
where $X_q$, $W_q$, $b_q$, and $y_q$ are the quantized versions of $X$, $W$, $b$, and $y$, respectively, $s_X$, $s_W$, $s_b$, and $s_y$ are the scales, while $z_X$, $z_W$, $z_b$, and $z_y$ are the zero points.
Even here there are four terms that are constant during inference:
\begin{equation}
    z_y + \frac{s_b}{s_y} (b_q - z_b) \qquad
    \frac{s_X s_W}{s_y} \qquad
    z_X \sum_{i=1}^{m} \sum_{j=1}^{n} W_{q,i,j} \qquad
    m~n~z_X z_W
\end{equation}

The same routine for view extraction of Conv2D is used also by the DepthwiseConv2D operator.

\subsection{AveragePool2D}

The AveragePool2D operator is typically used to downsample the input data by partitioning it into non-overlapping regions and computing the average value within each of them.
AveragePool2D works on 4D tensors with a single batch of matrices.
It performs average pooling on a per-channel basis, which means that the input channels are preserved throughout the pooling operation until the output is generated.
In this case, the operator quantization starts from the following output $y \in \mathbb{R}$, where $X \in \mathbb{R}^{m \times n}$ represents an input region, for a given channel:
\begin{equation}
    y = \frac{1}{m n} \sum_{i=1}^{m} \sum_{j=1}^{n} X_{i,j}
\end{equation}
The quantized output can be derived to obtain the following expression (details in Appendix~\ref{appendix:averagepool2d}):
\begin{equation}
\label{eq:averagepool2d}
    y_q = z_y + \frac{s_X}{s_y} \Bigg[ \bigg( \frac{1}{m n} \sum_{i=1}^{m} \sum_{j=1}^{n} X_{q,i,j}  \bigg) - z_X \Bigg]
\end{equation}
where $X_q$ and $y_q$ are the quantized versions of $X$ and $y$, respectively, $s_X$ and $s_y$ are the scales, while $z_X$ and $z_y$ are the zero points.
Even in this case there are a couple of constant terms that can be pre-computed:
\begin{equation}
    \frac{s_X}{s_y} \qquad
    \frac{1}{m n}
\end{equation}

Since the pooling operation is performed on an input region, the implementation of the AveragePool2D operator utilizes the same view extraction algorithm previously discussed.
Similarly to DepthwiseConv2D, the channels of the input region are not merged together.
Instead, the channel dimension is preserved.

\subsection{Activation Functions}

Activation functions transform the outputs of individual neurons to enable complex and expressive mappings between inputs and outputs.
They can be either applied as a separate operation after a specific layer or combined with an operator, in the latter case taking the name of \emph{fused} activation functions.
All the operators described so far support the addition of a fused activation function at the end of each iteration.
The activation functions currently implemented in MicroFlow are ReLU, ReLU6, and Softmax.
Their kernels remain the same, regardless of the application, and they are not pre-processed, since they contain little to no constant terms.
However, they have to be quantized, as explained next.

\subsubsection*{ReLU}

The ReLU function returns the input value if this is positive, or zero otherwise.
Its quantized expression is as follows (details in Appendix~\ref{appendix:relu}):
\begin{equation}
\label{eq:relu}
    y_q = 
    \begin{cases}
        z_y & \text{if $x_q < z_x$} \\
        z_y + \frac{s_x}{s_y} (x_q - z_x) & \text{if $x_q \geq z_x$} \\
    \end{cases}
\end{equation}
where $x_q$ and $y_q$ are the quantized versions of $x$ and $y$, respectively, $s_x$ and $s_y$ are the scales, while $z_x$ and $z_y$ are the zero points.
If ReLU is used as a fused activation function, then $s_x = s_y$ and $z_x = z_y$, and the previous expression becomes simply a $\max$ operator:
\begin{equation}
    y_q = \max(x_q, z)
\end{equation}

\subsubsection*{ReLU6}

The ReLU6 activation function is a variant of the standard ReLU that adds an upper bound constraint (i.e. the maximum activation value is $6$).
It is commonly used in applications where output values need to be limited to a specific range.
Due to its similarity to the ReLU function, the quantized version of ReLU6 can be immediately derived as follows:
\begin{equation}
\label{eq:relu6}
    y_q = 
    \begin{cases}
        \text{ReLU}(x_q, s_x, s_y, z_x, z_y) & \text{if $x_q < z_x + \frac{6}{s_x}$} \\
        z_y + \frac{6}{s_y} & \text{if $x_q \geq z_x + \frac{6}{s_x}$} \\
    \end{cases}
\end{equation}
where $x_q$ and $y_q$ are the quantized versions of the input $x \in \mathbb{R}$ and the output $y \in \mathbb{R}$, respectively, $s_x$ and $s_y$ are the scales, while $z_x$ and $z_y$ are the zero points.
If ReLU6 is used as a fused activation function, then $s_x = s_y$ and $z_x = z_y$, resulting in the following expression:
\begin{equation}
    y_q = \min\left(\max(x_q, z), z + \frac{6}{s}\right)
\end{equation}

\subsubsection*{Softmax}

The last activation function is Softmax, which is commonly used in deep learning for multi-class classification problems.
It takes a vector of real-valued scores as input and transforms them into a probability distribution over multiple classes. 
Its quantized expression is as follows (details in Appendix~\ref{appendix:softmax}):
\begin{equation}
\label{eq:softmax}
    y_{q,i} = z_y + \frac{e^{s_x x_{q,i}}}{s_y \sum_{j=1}^{n} e^{s_x x_{q,j}}}
\end{equation}
where $x_q$ and $y_q$ are the quantized versions of the input $x \in \mathbb{R}$ and the output $y \in \mathbb{R}$, respectively, $s_x$ and $s_y$ are the scales, while $z_x$ and $z_y$ are the zero points.

\section{Experimental Evaluation}
\label{sec:evaluation}

To evaluate the performance of the proposed solutions, several models of increasing complexity have been implemented in MicroFlow and compared against the state-of-the-art, analysing in particular their accuracy, memory footprint, inference time, and energy consumption. The experimental setup and the results are presented next, followed by a brief discussion of the main outcomes and some further insights.

\subsection{Setup} \label{sec:setup}

MicroFlow has been evaluated on the three models listed in Table~\ref{tab:models-summary}, quantized to 8-bit signed integers, and datasets with different size and complexity, often used in the literature to evaluate TinyML systems~\cite{Zhang2022}:
\begin{itemize}
    \item a simple sine predictor~\cite{TFLMSine2024};
    \item a speech command recognizer~\cite{TFLMSpeech2024};
    \item a person detector~\cite{TFLMPerson2024}.
\end{itemize}

\begin{table}
	\centering
	\def\arraystretch{1.2}
 \scriptsize
	\begin{tabular}{|p{2.5cm}|l|p{2.2cm}|l|l|l|p{1.8cm}|}
		\hline
		\textbf{Model} & \textbf{Architecture} & \textbf{Operators} & \textbf{Layers} & \textbf{Size} & \textbf{Dataset} & \textbf{Data Type} \\ \hline
		Sine predictor~\cite{TFLMSine2024} & Custom NN & FullyConnected \newline ReLU & 3 & 3kB & Custom sine wave & Floating-point \newline (FP32) \\ \hline
		Speech command \newline recognizer~\cite{TFLMSpeech2024} & TinyConv~\cite{TinyConv} & DepthwiseConv2D \newline FullyConnected \newline Softmax & 4 & 19kB & Speech Command v2~\cite{warden2018speech} & Audio  \\ \hline
		Person detector~\cite{TFLMPerson2024}& MobileNet~\cite{howard2017mobilenets} & DepthwiseConv2D \newline Conv2d \newline AveragePool2D \newline Softmax & 30 & 301kB & Visual Wake Words~\cite{chowdhery2019visual} & Image\\ \hline
	\end{tabular}
	\caption{Summary of models and datasets used for the evaluation of MicroFlow, as described in Sec.~\ref{sec:setup}. They are ordered based on increasing deployment cost, which depends mostly on number of Layers and memory Size. Since quantized weights are int8 (byte) encoded, note that Size reflects also the number of the model's parameters.} \label{tab:models-summary}
	
\end{table}

\begin{figure}
    \centering
    \includegraphics[width=4.3cm]{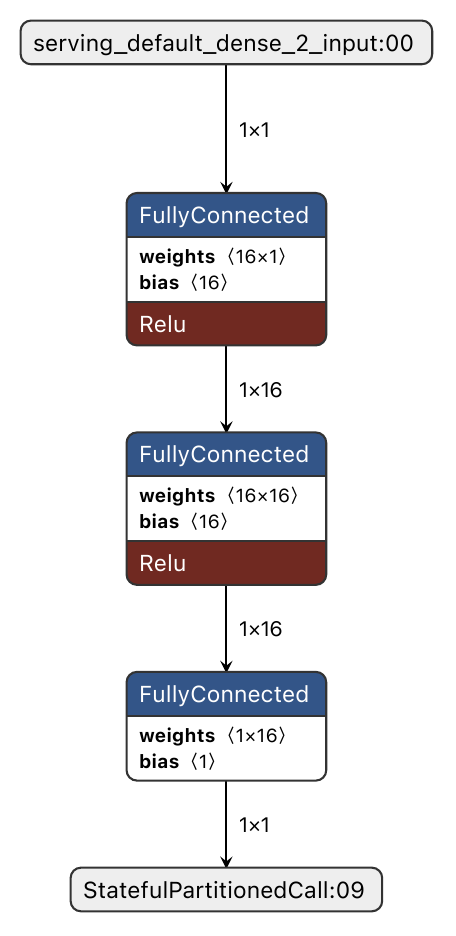}
    \hfill
    \includegraphics[width=2.4cm]{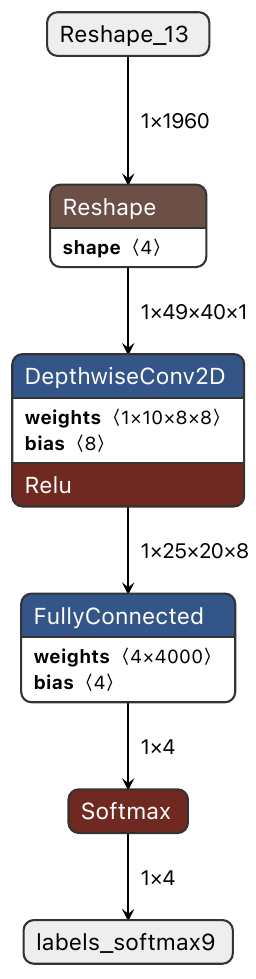}
    \hfill
    \includegraphics[width=5.4cm]{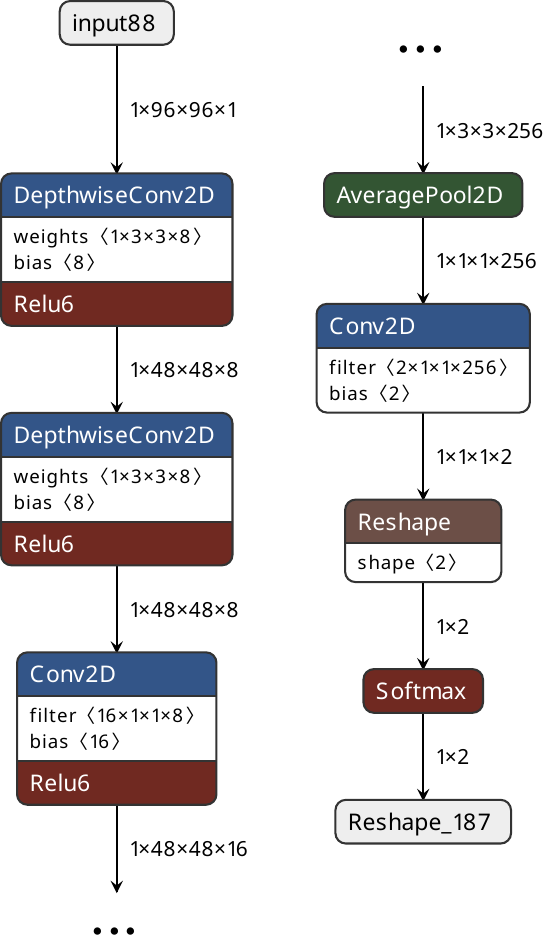}
    \caption{Schematic representation of the sine predictor model (left)~\cite{TFLMSine2024}, the speech command recognizer model (centre)~\cite{TFLMSpeech2024}, and the person detector model (right)~\cite{TFLMPerson2024}. In the latter, the intermediate repeated pattern of layers is omitted for simplicity.}
    \label{fig:models}
\end{figure}

The sine predictor is the simplest and smallest model, which was chosen to evaluate the systems with extremely minimal computational and memory resources, typical of 8-bit MCUs.
As the name suggests, given an input $x \in \mathbb{R}$, the sine model returns an output $y = \sin(x)$.
The pre-trained model used in this work is available on the TFLM repository~\cite{TFLMSine2024}. It was evaluated on 1000 testing samples of floating-point~(FP32) values, generated by a $\sin(\cdot)$ function with some random uniform noise $n \sim \mathcal{U}(-0.1, 0.1)$.
As shown in Fig.~\ref{fig:models}~(left), the model includes three FullyConnected layers with 16 neurons each, the first two using a ReLU fused activation function.
The model's size is approximately 3kB.

The goal of the second model is to recognize two spoken words, \emph{yes} and \emph{no}, resembling the wake-word detection task used in many real-world applications.
It incorporates both convolutional operations and dense layers, providing a good test-bed for evaluation purposes.
In particular, the model uses convolutional operations applied to an input signal, which is the Fast Fourier Transform~(FFT) of an audio sample, and outputs the likelihood for the input sample to be in one of these categories: the \emph{yes} word, the \emph{no} word, silence, or unknown.
It is a \emph{TinyConv} architecture~\cite{TinyConv}, employing a DepthwiseConv2D layer followed by a FullyConnected one.
The output scores are then converted to probabilities by a Softmax activation layer.
Its schematic structure is illustrated in Fig.~\ref{fig:models}~(centre).
Since the speech command recognizer works with convolutional layers, the tensors are 4-dimensional, and the overall size reaches approximately 19kB. The model used in this work is available on the TFLM repository~\cite{TFLMSpeech2024}. It was pre-trained on the \emph{Speech Commands Dataset} v.2~\cite{warden2018speech}, and then evaluated on its own 1236 testing samples.

The person detector is the largest and most complex model of the three.
Given an input frame, its task is to detect the presence of a person.
In particular, the model takes a grayscale image of $96 \times 96$ pixels and outputs the probabilities of the two classes \emph{person} and \emph{not-person}.
The architecture is based on \emph{MobileNet} version~1~\cite{howard2017mobilenets} and employs a series of DepthwiseConv2D operators followed by Conv2D operators.
The end of the chain consists of an AveragePool2D layer, followed by a Conv2D operator, and a final Softmax activation function layer.
A simplified representation of the model is shown in Fig.~\ref{fig:models}~(right). Similarly to the speech command recognizer, this is available on the TFLM repository~\cite{TFLMPerson2024}. It was pre-trained on the \emph{Visual Wake Words Dataset}~\cite{chowdhery2019visual}, and then evaluated on its own 406 testing samples. 
Due to the type and quantity of operators involved, which require 4D tensors, the size of the model reaches 301kB.
For this reason, in the following experiments, the person detector model will be evaluated only on MCUs with sufficient Flash memory to accommodate its size.

Although MicroFlow can run on a variety of systems, the analysis here focuses on bare-metal embedded devices.
Specifically, the evaluation was done on the MCU and test boards detailed in see Table~\ref{tab:hardware-summary}, listed in descending order based on their performance and considering resource constraints:
\begin{itemize}
    \item ESP32 (Adafruit HUZZAH32);
    \item ATSAMV71 (SAM V71 Xplained Ultra);
    \item nRF52840 (Arduino Nano 33 BLE Sense);
    \item LM3S6965 (emulated by QEMU\footnote{\url{https://www.qemu.org/}});
    \item ATmega328 (Arduino Uno).
\end{itemize}
This choice covers a wide range of possible memory sizes, architectures, and peripherals.
From the high-performance 32-bit ESP32 with 4MB of Flash and 328kB of RAM, to the 8-bit ATmega328, with only 32kB of Flash and 2kB of RAM.

\begin{table}
	\centering
	\def\arraystretch{1.2}
 \small
	\begin{tabular}{|l|l|l|l|l|}
		\hline
		\textbf{MCU} & \textbf{Architecture} & \textbf{Flash Size}  & \textbf{RAM Size} & \textbf{Clock} \\ \hline
		ESP32             & 32-bit Xtensa                  & 4MB   & 328kB   & 240MHz   \\ \hline
		ATSAMV71          & 32-bit Cortex-M7F              & 2MB  & 384kB   & 300MHz   \\ \hline
		nRF52840          & 32-bit Cortex-M4F              & 1MB  & 256kB   & 64MHz    \\ \hline
		LM3S6965          & 32-bit Cortex-M3               & 256kB & 64kB    & 50MHz    \\ \hline
		ATmega328         & 8-bit AVR                      & 32kB  & 2kB     & 20MHz    \\ \hline
	\end{tabular}
	\caption{Specifications of the MCUs used for the experiments.}
	\label{tab:hardware-summary}
\end{table}

\subsection{Experiments}

MicroFlow was evaluated against TFLM~\cite{MLSYS2021_d2ddea18}, 
since the latter provides state-of-the-art performance and it is widely adopted in the TinyML community.
The two frameworks share some similarities in terms of software architecture and target applications.
However, to the best of our knowledge, there are no previous studies that compare TinyML models using Rust, as in MicroFlow, to other C/C++ frameworks, like TFLM.
The following experiments cover some of the most significant TinyML objectives, namely:
\begin{itemize}
    \item accuracy;
    \item memory usage;
    \item runtime performance;
    \item energy consumption.
\end{itemize}

The sine predictor was assessed by measuring the Mean Squared Error~(MSE) and the Root Mean Squared Error~(RMSE), while Precision, Recall, and \textit{F}$_1$ Score were computed for the other two NN models.
For the speech command recognizer, which has four output classes, these metrics have been averaged to provide an overall accuracy across all of them.
The resulting metrics are then compared between the MicroFlow and the TFLM inference engines. The experimental results are detailed in Sec.~\ref{sec:accuracy}.

The memory experiments, presented next in Sec.~\ref{sec:memory_usage}, considered both Flash and RAM usage,
loading a minimal firmware for model inference and analyzing the compiled binary.
This minimal firmware was created avoiding any platform-dependent factors that could compromise a fair evaluation (e.g., it does not include printing statements, because their implementation varies across platforms and architectures, generating biased results).

The assessment of the runtime performance compared the inference execution times of the three NN models on different MCUs. Specifically, the execution times were measured by the MCU timers for 100 iterations, then the median times were computed and evaluated. The results are presented in Sec.~\ref{sec:runtime}.

Finally, the last experiments estimated and compared the energy consumption of MircoFlow and TLFM on different architectures.
Knowing the execution times and measuring the average power usage of the MCUs, it was possible to compute the average energy consumption during the inference process with the three models.
Details and results of these experiments are discussed in Sec.~\ref{sec:energy}.

\subsubsection{Accuracy}
\label{sec:accuracy}

The results of the accuracy experiments are shown in Table~\ref{tab:accuracy-results}.
For the sine model, $1000$ samples were randomly generated by adding some noise -- uniformly distributed in $[-0.1,0.1]$ -- to the original sine function, and the MSE was calculated against the actual values of the function.
The test sets used for the other two models were the ones provided by the original training datasets, namely the Speech Commands and the Visual Wake Words datasets mentioned in Sec.~\ref{sec:setup}.
The results show that the two inference engines performed very similarly.
This demonstrates that MicroFlow's operators were correctly computed and implemented, enabling the achievement of state-of-the-art classification performance on TinyML architectures.
In particular, the results show a very low error achieved by both the TFLM and the MicroFlow's implementations of the sine predictor. Also, the overall performance for speech command recognition is obviously better than person detection, due to the increased difficulty of the latter. For both models though, TFLM and MicroFlow are still on par.

\begin{table}
	\centering
	\def\arraystretch{1.2}
        \small
	\begin{tabular}{|l|l|l|}
		\multicolumn{3}{c}{\textbf{Sine Predictor}}            \\ \hline
		             & \textbf{TFLM} & \textbf{MicroFlow} \\ \hline
		\textbf{MSE}     & 0.0157             & 0.0154             \\ \hline
		\textbf{RMSE}     & 0.1253             & 0.1241             \\ \hline
	\end{tabular}
        \hfil
	\begin{tabular}{|l|l|l|}
		\multicolumn{3}{c}{\textbf{Speech Command Recognizer}} \\ \hline
		             & \textbf{TFLM} & \textbf{MicroFlow} \\ \hline
		\textbf{Precision}    & $91.737\%$         & $91.638\%$         \\ \hline
        \textbf{Recall}       & $88.611\%$         & $88.972\%$         \\ \hline
        \textbf{\textit{F}$_1$ Score}  & $90.147\%$         & $90.285\%$         \\ \hline
	\end{tabular}
	\begin{tabular}{|l|l|}
		\multicolumn{2}{c}{\textbf{Person Detector}}           \\ \hline
		             \textbf{TFLM} & \textbf{MicroFlow} \\ \hline
		$71.843\%$         & $72.003\%$         \\ \hline
        $85.382\%$         & $85.401\%$         \\ \hline
        $78.030\%$         & $78.132\%$         \\ \hline
	\end{tabular}
	\caption{Results of the accuracy experiment performed on the three inference models. The first one~(left) is a simple sine wave with some additive uniform noise, for which Mean Squared Error~(MSE) and Root Mean Squared Error~(RMSE) are calculated against the actual values of the function. The second and the third ones are two classifiers for speech command recognition~(middle) and person detection~(right), respectively, the accuracy of which are measured in terms of Precision, Recall, and \textit{F}$_1$ Score.
    \label{tab:accuracy-results}}
\end{table}

Since the original NNs were the same, one would expect the results of these experiments to be identical as well. However, the small differences between MicroFlow and TFLM's accuracy can be explained 
by analyzing the intermediate quantized outputs of the NNs, and noticing that occasionally the output tensors of some operators differ by one integer unit (sometimes TFLM was one unit above, sometimes one unit below).
This is caused to small differences in the floating-point implementations of the two architectures, which led to rounding discrepancies.
Also, since MicroFlow and TFLM are based on two different programming languages and compilers, variations in the latter's built-in operations could explain the observed results.

\subsubsection{Memory Usage}
\label{sec:memory_usage}

\begin{figure}[b]
    \centering
    \includegraphics[width=.55\textwidth]{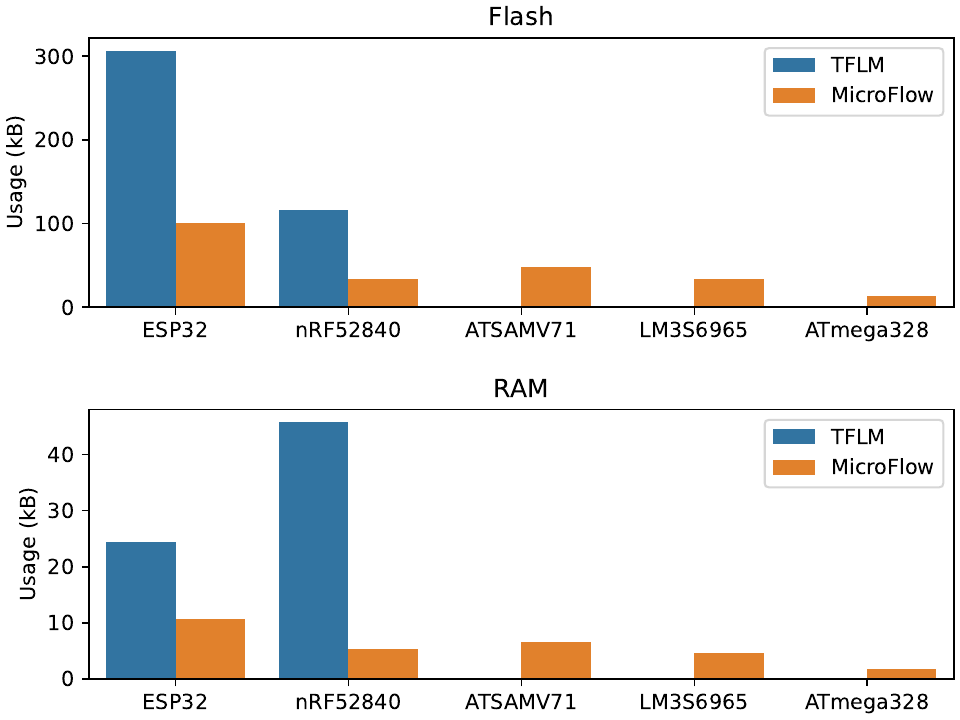}
    \caption{Results of the memory usage experiment for the sine predictor model. The results show the Flash (top) and RAM~(bottom) memory used, in kB, by the binaries compiled with TFLM and MicroFlow, for each MCU under consideration. \label{fig:sine-memory}}
\end{figure}

For the sine model, the results of the experiment can be observed in Fig.~\ref{fig:sine-memory}.
The chart shows the Flash and maximum RAM memory usage by the compiled binary for each MCU.
As expected, there are significant differences between the two inference engines. For example, on the ESP32, MicroFlow uses $\sim65\%$ less Flash memory than TFLM; on the nRF52840, it uses only 5.296kB of RAM, against the 45.728kB required by TFLM.
It should be noted that the very low memory usage of MicroFlow made it possible to successfully build and run the sine model on all the tested MCUs.
In particular, it was possible to perform inference even on the 8-bit AVR ATmega328, with a small Flash and RAM footprint of 13.619kB and 1.706kB, respectively.
Since it requires significantly more memory, TFLM could run only on the ESP32 and the nRF52840 MCUs

The results' difference was expected since TFLM is an interpreter-based engine that needs to be loaded on the MCU regardless of the model's size, which is not known at compile-time.
Because of this, all the operator kernels must be loaded too, occupying further memory space.
MicroFlow instead loads only the necessary weights and operator kernels, chosen at compile-time, leaving out the model's parts that are not needed at runtime (e.g. operator versions, tensor names, sizes, options, etc.).
For example, instead of storing all the kernel versions of an operator and choose one at runtime based on the model, MicroFlow reads the operator version at compile-time and stores only this in the target memory.

\begin{figure}
    \centering
    \includegraphics[width=.47\textwidth]{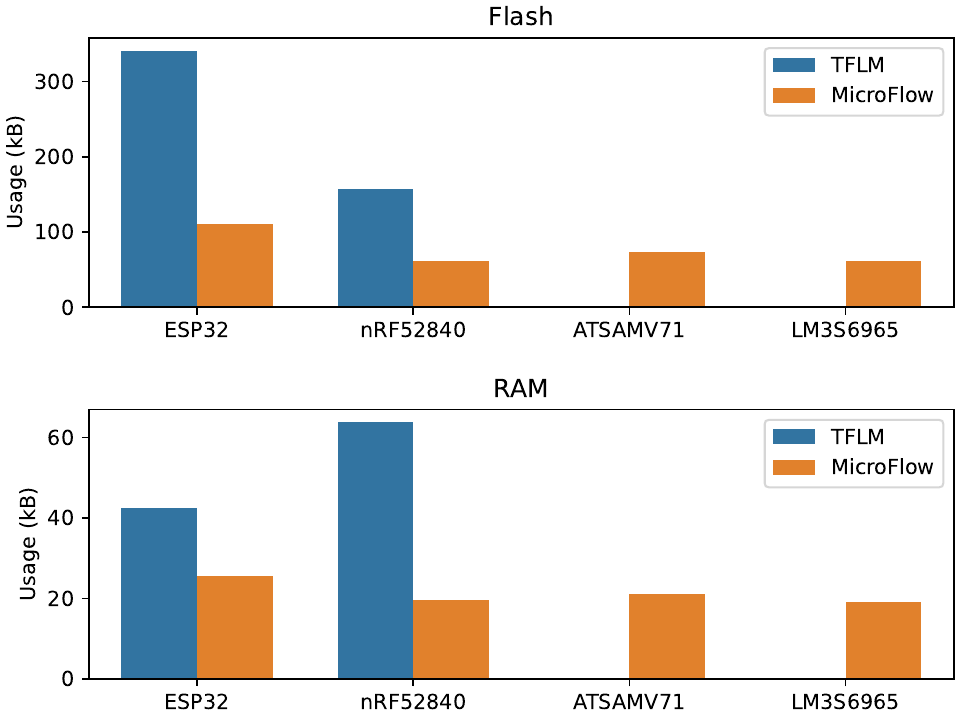} \hfill
    \includegraphics[width=.47\textwidth]{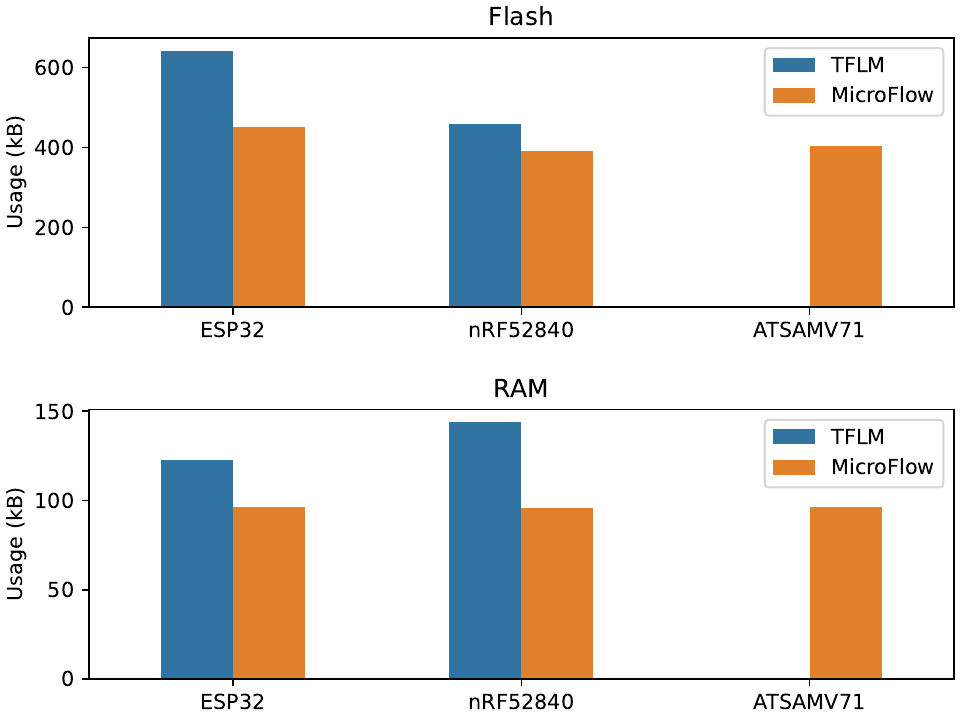}
    \caption{Results of the memory usage experiment for the speech command recognizer (left) and person detector (right). The figures show the Flash (top) and RAM~(bottom) memory used, in kB, by the binaries compiled with TFLM and MicroFlow, for each MCU under consideration. \label{fig:memory}}
\end{figure}

Similar considerations can be done for the experiments with the other models. In particular, the results for the speech command recognizer and person detector can be seen in Fig.~\ref{fig:memory}.
Even here, only the most powerful MCUs (ESP32 and nRF52840) were used for TFLM because the model's weights would simply not fit in the memory of the remaining ones.
In addition, the ATmega328 and LM3S6965 were progressively excluded from the speech command recognition and person detection experiments, respectively, due to the models being too large for these architectures too.

The results in Fig.~\ref{fig:memory} demonstrate that MicroFlow consistently outperforms TFLM in terms of memory efficiency, requiring significantly less Flash and RAM to run models on several architectures, even on those more constrained than the ESP32 and nRF52840.
However, as the model's size increases, the gap between MicroFlow and TFLM becomes smaller, since the NN weights occupy most of the allocated memory.
The impact of the interpreter overhead is therefore smaller but still noticeable.
Indeed, even on the more complex person detector, MicroFlow still saved more than $15\%$ of memory compared to TFLM.

\subsubsection{Runtime Performance}
\label{sec:runtime}

\begin{figure}[t]
    \centering
    \includegraphics[width=.6\textwidth]{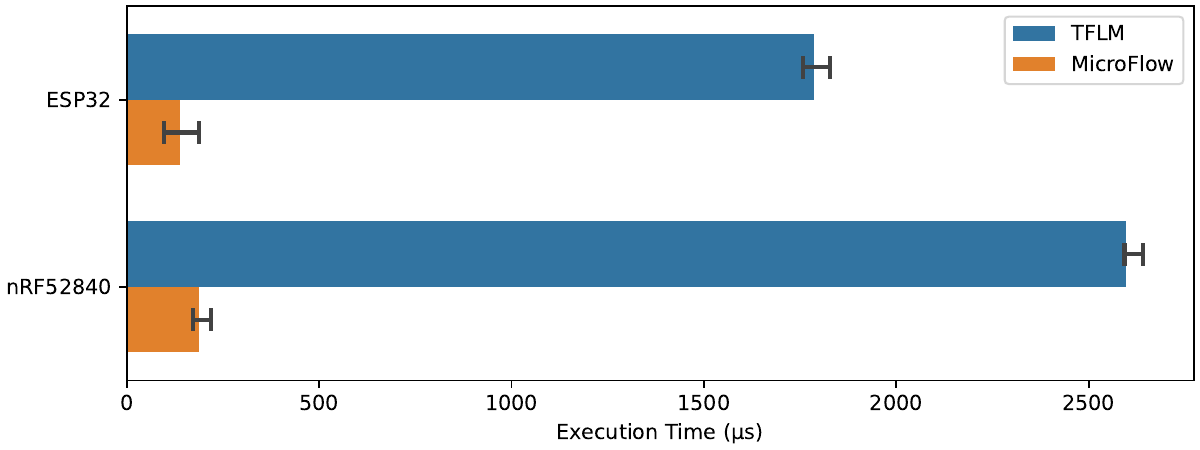}
    \includegraphics[width=.6\textwidth]{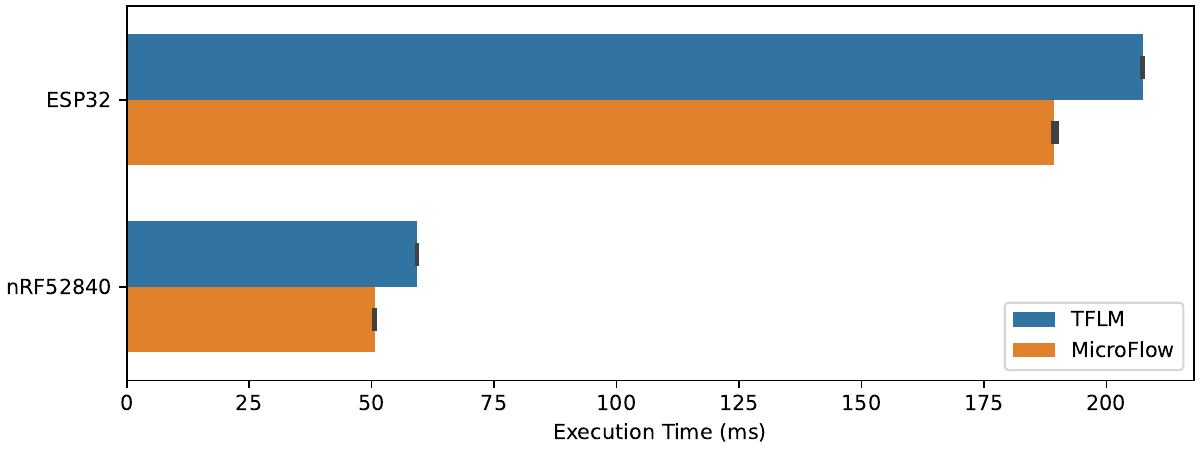} \hfill
    \includegraphics[width=.6\textwidth]{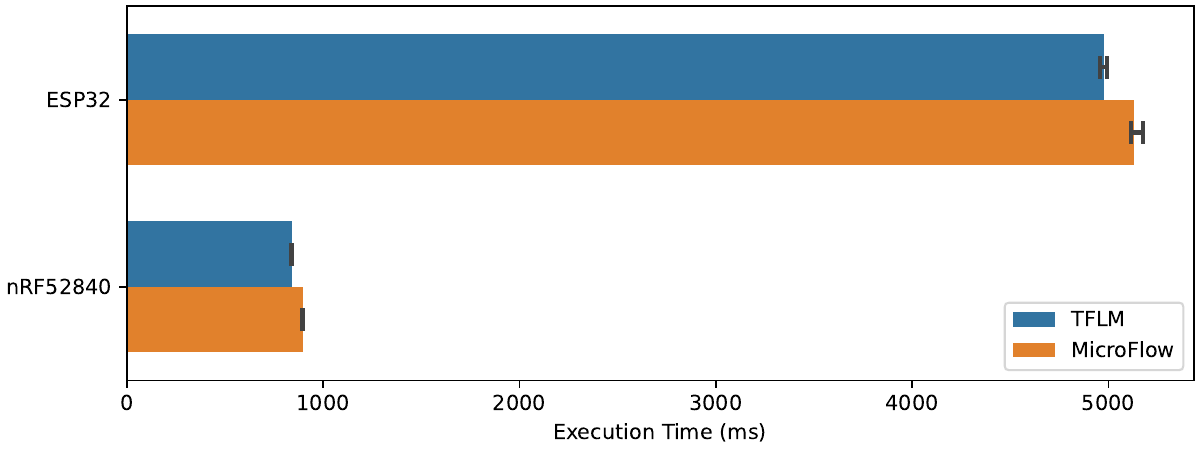}
    \caption{Runtime results for sine predictor~(top), speech command recognizer~(middle), and person detector~(bottom), comparing MicroFlow and TFLM's inference times, in ms, of the three models on the two MCUs supported by both frameworks.}
    \label{fig:runtime-comparison}
\end{figure}

The next experiments compared the actual inference times of MicroFlow and TFLM. They were therefore performed only on the MCUs supported by both frameworks, namely the ESP32 and the nRF52840.
The firmware used consists of a minimal program that cyclically performed inference on a given model for 100 iterations, using the MCU's timers to measure the execution time of each cycle.
The plots in Fig.~\ref{fig:runtime-comparison} present the median results with the $95\%$ percentile interval.

For the sine predictor, the plot shows that MicroFlow was ten times faster than TFLM on both the MCUs.
The gain can be explained by two main factors:
first, the interpreter of TFLM introduces an overhead during the inference process due to the interpretation of the model's operations, the dynamic memory management, and other interpreter-related tasks
(this is even more pronounced when the model is small, since the inference execution time is comparable to the interpreter overhead);
second, MicroFlow exploits the efficiency of the Rust programming language and the substantial pre-processing phase that, thanks to the compiler's static analysis and optimizations, results in less CPU operations and overall execution time.

However, the performance gap between MicroFlow and TFLM is narrower in the next experiments, since the increased size and complexity of the models make the actual inference process more significant than the interpreter overhead.
Indeed, for the speech command recognizer, Fig.~\ref{fig:runtime-comparison} shows less differences between the two inference engines, although MicroFlow's performance is still $9\%$ better on the ESP32 and $15\%$ better on the nRF52840 compared to TFLM.
It is also interesting to observe that
the nRF52840 was more than three times faster than the ESP32.
This might seem counterintuitive, since the ESP32's CPU clock is significantly faster than the nRF52840's.
However, the Floating-Point Unit~(FPU) of the former is known to be not very efficient~\cite{ESP32FP}, which could negatively impact on the performance of the inference engine.

Finally, the outcome for the person detector experiment, also illustrated in Fig.~\ref{fig:runtime-comparison}, shows that TFLM performed slightly better than MicroFlow, although the gap is relatively small, close to $6\%$.
The result stems from the MobileNet model of the person detector, which consists mostly of convolutional operations.
As the size of the model increases, the execution time is dominated by these computationally intensive operations, reaching a saturation point that depends on the hardware constraints of the MCU (i.e. CPU clock speed and number of cores).
Unfortunately, little can be done in terms of software optimizations.
The slightly faster execution of TFLM is achieved thanks to the optimized kernels provided by the MCU manufacturer\footnote{For the nRF52840, these kernels are contained in the \emph{CMSIS-NN} software library developed by ARM.}, currently unavailable on MicroFlow.
As in the previous experiment, the ESP32 performed significantly worse than the nRF52840 due to the FPU.

Overall, these results demonstrate a significant advantage of MicroFlow compared to TFLM in terms of inference time, at least for models that cannot benefit from hardware optimizations. Such an advantage disappears when the interpreter overhead of TFLM is outweighed by the computational complexity of operations like dense convolution layers, which would require further hardware and software optimization techniques~\cite{Chen2020,Hacene2020}.

\subsubsection{Energy Consumption}
\label{sec:energy}

The total energy consumption (average power by execution time) of the last experiment is reported in Table~\ref{tab:energy-results}.
As expected, the consumption increases with the complexity of the models and is affected by the specific embedded system's architecture~\cite{Reddi2020,inference-tiny-12}. The table shows that the simple sine predictor requires very little energy, while the other two models exhibit significantly higher energy usage. Additionally, the energy consumption of the latter varies considerably between ESP32 and nRF52840.

Note that, since the average power usage (per operation) of MicroFlow and TFLM is similar, the final energy consumption values are proportional to the execution times.
This behavior indeed can be attributed to the fact that the types of operations performed by the MCU are essentially the same for the two inference engines -- hence similar power usage.
Moreover, the peripherals of the MCU used by the inference engines are identical, resulting in limited possibilities for power optimizations.
Since the energy consumption of the two inference engines is directly proportional to their execution times, MicroFlow's models are generally more energy-efficient than, or at least comparable to, TFLM's, as confirmed by the results in Table~\ref{tab:energy-results}.

\begin{table}
	\centering
	\def\arraystretch{1.2}
 \small
	\begin{tabular}{|l|l|l|}
		\multicolumn{3}{c}{\textbf{Sine Predictor}}                              \\ \hline
		         & \textbf{TFLM}            & \textbf{MicroFlow}            \\ \hline
		\textbf{ESP32}    & 149nWh     & 11nWh      \\ \hline
		\textbf{nRF52840} & 216nWh     & 16nWh      \\ \hline
	\end{tabular}
 	\begin{tabular}{|l|l|}
		\multicolumn{2}{c}{\textbf{Speech Command Recognizer}}                   \\ \hline
		         \textbf{TFLM}            & \textbf{MicroFlow}            \\ \hline
		23.05mWh  & 21.04mWh  \\ \hline
		6.58mWh   & 5.62mWh   \\ \hline
	\end{tabular}
 	\begin{tabular}{|l|l|}
		\multicolumn{2}{c}{\textbf{Person Detector}}                             \\ \hline
		         \textbf{TFLM}            & \textbf{MicroFlow}            \\ \hline
		691.11mWh & 694.44mWh \\ \hline
		116.58mWh & 124.44mWh \\ \hline
	\end{tabular}
	\caption{Results of the comparison between MicroFlow and TFLM's energy consumption, in nWh for the sine predictor~(left), and in mWh for the speech command recognizer~(middle) and the person detector~(right). The results are only for the two MCUs supported by both frameworks.\label{tab:energy-results}}
\end{table}

\subsection{Discussion}
\label{sec:discussion}
MicroFlow targets mainly low-cost and low-energy embedded systems.
To this end, it has been evaluated on several MCUs with limited resources, some of which imposed severe constraints on the model's size and performance.
Among these, the most prominent one is memory size. For example, any attempt to flash the Person Detector binary on the ATmega328 resulted in a ``not enough memory'' error.
Nevertheless, the framework is suitable for any platforms capable of running Rust, therefore it can be deployed on more powerful embedded systems to perform runtime inference on complex models.
Indeed, with sufficient computational power for compilation and adequate resources for inference, running larger models presents no evident challenges.
However, it was noticed that the performance of MicroFlow can reach a saturation point with larger and more complex models. Although this can be explained by the current lack of hardware-specific optimizations, further experiments are neded to determine whether an upper-bound exists with respect to TFLM or other MCU-tailored frameworks.

\section{Conclusions}
\label{sec:conclusions}

This paper presented MicroFlow, a TinyML inference engine for resource-constrained MCUs, which is implemented in Rust to achieve memory safety and efficiency. Experimental results demonstrated its effectiveness and excellent performance in several ML tasks, balancing state-of-the-art accuracy with efficient memory usage, execution time, and energy consumption.
In particular, it outperformed current TinyML standards in terms of memory requirements, while achieving better or, in the worst case, similar execution time results.
MicroFlow has been released as an open-source project for the benefit of the research and industrial communities, with the possibility to expand its functionalities according to the needs of the final user, thanks to its modular design.

As in every project, there is always room for improvement.
Future version of the software, which is actively maintained and already used in real-world applications\footnote{\url{https://www.grepit.se/}}, will include new operators.
This in turn will allow MicroFlow to support a wider range of operations and NN architectures.
Moreover, possible optimizations techniques could be investigated that exploit specific hardware features (e.g. NPU and other AI accelerators) of the particular MCU family.
Further experiments to assess the benefits of these optimizations, particularly as the model size increases, would offer valuable insights into the competitive advantage and the true scalability of the proposed framework.
Finally, promising areas for future research include expanding MicroFlow's capabilities to enable further scientific advancements on trending TinyML topics such as on-device learning, continual learning, and federated learning~\cite{tinyml-continual-2021, tinyml-federated-2024}.

\appendix

\section{Operators Details}
\label{appendix:operators}

\subsection{FullyConnected Operator} \label{appendix:fullyconnected}

\begin{equation}
\begin{split}
    Y_{i,j}
    &= s_b(b_{q,j} - z_b) + \sum_{k=1}^{n} s_X(X_{q,i,k} - z_X) s_W(W_{q,k,j} - z_W) \\
    &= s_b(b_{q,j} - z_b) + s_X s_W \sum_{k=1}^{n} (X_{q,i,k} - z_X) (W_{q,k,j} - z_W) \\
    &= s_b(b_{q,j} - z_b) + s_X s_W \Bigg[ \bigg( \sum_{k=1}^{n} X_{q,i,k} W_{q,k,j} \bigg) - \bigg( z_W \sum_{k=1}^{n} X_{q,i,k} \bigg) \\
    & \qquad - \bigg( z_X \sum_{k=1}^{n} W_{q,k,j} \bigg) + n~z_X z_W \Bigg] \\
    &= s_Y(Y_{q,i,j} - z_Y)
\end{split}
\end{equation}
where $X_q$, $W_q$, $b_q$, and $Y_q$ are the quantized versions of $X$, $W$, $b$, and $Y$, respectively, $s_X$, $s_W$, $s_b$, and $s_Y$ are the scales, while $z_X$, $z_W$, $z_b$, and $z_Y$ are the zero points.
Therefore:
\begin{equation}
\begin{split}
    Y_{q,i,j} &= z_Y + \frac{s_b}{s_Y} (b_{q,j} - z_b) + \frac{s_X s_W}{s_Y} \Bigg[ \bigg( \sum_{k=1}^{n} X_{q,i,k} W_{q,k,j} \bigg) - \bigg( z_W \sum_{k=1}^{n} X_{q,i,k} \bigg) \\
    & \qquad - \bigg( z_X \sum_{k=1}^{n} W_{q,k,j} \bigg) + n~z_X z_W \Bigg]
\end{split}
\end{equation}

\subsection{Conv2D Operator} \label{appendix:conv2d}

\begin{equation}
\begin{split}
    y
    &= s_b(b_q - z_b) + \sum_{i=1}^{m} \sum_{j=1}^{n} \sum_{k=1}^{c} s_X(X_{q,i,j,k} - z_X) s_F(F_{q,i,j,k} - z_F) \\
    &= s_b(b_q - z_b) + s_X s_F \sum_{i=1}^{m} \sum_{j=1}^{n} \sum_{k=1}^{c} (X_{q,i,j,k} - z_X) (F_{q,i,j,k} - z_F) \\
    &= s_b(b_q - z_b) + s_X s_F \Bigg[ \bigg( \sum_{i=1}^{m} \sum_{j=1}^{n} \sum_{k=1}^{c} X_{q,i,j,k} F_{q,i,j,k} \bigg) - \bigg( z_F \sum_{i=1}^{m} \sum_{j=1}^{n} \sum_{k=1}^{c} X_{q,i,j,k} \bigg) \\
    & \qquad - \bigg( z_X \sum_{i=1}^{m} \sum_{j=1}^{n} \sum_{k=1}^{c} F_{q,i,j,k} \bigg) + m~n~c~z_X z_F \Bigg] \\
    &= s_y(y_q - z_y)
\end{split}
\end{equation}
where $X_q$, $F_q$, $b_q$, and $y_q$ are the quantized versions of $X$, $F$, $b$, and $y$, respectively, $s_X$, $s_F$, $s_b$, and $s_y$ are the scales, while $z_X$, $z_F$, $z_b$, and $z_y$ are the zero points.
Therefore:
\begin{equation}
\begin{split}
    y_q &= z_y + \frac{s_b}{s_Y} (b_q - z_b) + \frac{s_X s_F}{s_y} \Bigg[ \bigg( \sum_{i=1}^{m} \sum_{j=1}^{n} \sum_{k=1}^{c} X_{q,i,j,k} F_{q,i,j,k} \bigg) \\
    & \qquad - \bigg( z_F \sum_{i=1}^{m} \sum_{j=1}^{n} \sum_{k=1}^{c} X_{q,i,j,k} \bigg) - \bigg( z_X \sum_{i=1}^{m} \sum_{j=1}^{n} \sum_{k=1}^{c} F_{q,i,j,k} \bigg) + m~n~c~z_X z_F \Bigg]
\end{split}
\end{equation}

The Conv2D operator requires also a view extraction routine in the kernel to select the appropriate input elements for each convolution, taking as arguments: input tensor, view dimensions, padding, and strides.
For each position in the input, the algorithm calculates the neighboring components to include in the view.

\begin{algorithm}
    \caption{View extraction algorithm for the Conv2D operator.}
    \label{alg:view-extraction}
    \begin{algorithmic}
        \REQUIRE $X \in \mathbb{R}^{m \times n}$
        \REQUIRE $V \in \mathbb{R}^{p \times q}$
        \REQUIRE $\text{padding} \in \{\text{Same}, \text{Valid}\}$
        \REQUIRE $\text{stride}_h, \text{stride}_w \in \mathbb{N}$

        \STATE $\text{shift}_h \gets \lfloor \frac{p - 1}{2} \rfloor$
        \STATE $\text{shift}_w \gets \lfloor \frac{q - 1}{2} \rfloor$

        \FOR{$i \in [0, m)$, $j \in [0, n)$, $k \in [0, p)$, $l \in [0, q)$}
            \STATE $\text{index}_h \gets \text{stride}_h * i + k$
            \STATE $\text{index}_w \gets \text{stride}_w * j + l$
            \IF{$\text{padding} = \text{Same}$}
                \STATE $\text{index}_h \gets \text{index}_h - \text{shift}_h$
                \STATE $\text{index}_w \gets \text{index}_w - \text{shift}_w$
                \IF{$\text{index}_h \in [0, m)$ \AND $\text{index}_w \in [0, n)$}
                    \STATE $V_{k, l} \gets X_{\text{index}_h, \text{index}_w}$
                \ELSE
                    \STATE $V_{k, l} \gets 0$
                \ENDIF
            \ELSIF{$\text{padding} = \text{Valid}$}
                \STATE $V_{k, l} \gets X_{\text{index}_h, \text{index}_w}$
            \ENDIF
        \ENDFOR
        \RETURN $V$
    \end{algorithmic}
\end{algorithm}

\subsection{DepthwiseConv2D} \label{appendix:depthwiseconv2d}

\begin{equation}
\begin{split}
    y
    &= s_b(b_q - z_b) + \sum_{i=1}^{m} \sum_{j=1}^{n} s_X(X_{q,i,j} - z_X) s_W(W_{q,i,j} - z_W) \\
    &= s_b(b_q - z_b) + s_X s_W \sum_{i=1}^{m} \sum_{j=1}^{n} (X_{q,i,j} - z_X) (W_{q,i,j} - z_W) \\
    &= s_b(b_q - z_b) + s_X s_W \Bigg[ \bigg( \sum_{i=1}^{m} \sum_{j=1}^{n} X_{q,i,j} W_{q,i,j} \bigg) - \bigg( z_W \sum_{i=1}^{m} \sum_{j=1}^{n} X_{q,i,j} \bigg) \\
    & \qquad - \bigg( z_X \sum_{i=1}^{m} \sum_{j=1}^{n} W_{q,i,j} \bigg) + m~n~z_X z_W \Bigg] \\
    &= s_y(y_q - z_y)
\end{split}
\end{equation}
where $X_q$, $W_q$, $b_q$, and $y_q$ are the quantized versions of $X$, $W$, $b$, and $y$, respectively, $s_X$, $s_W$, $s_b$, and $s_y$ are the scales, while $z_X$, $z_W$, $z_b$, and $z_y$ are the zero points.
Therefore:
\begin{equation}
\begin{split}
    y_q &= z_y + \frac{s_b}{s_y} (b_q - z_b) + \frac{s_X s_W}{s_y} \Bigg[ \bigg( \sum_{i=1}^{m} \sum_{j=1}^{n} X_{q,i,j} W_{q,i,j} \bigg) - \bigg( z_W \sum_{i=1}^{m} \sum_{j=1}^{n} X_{q,i,j} \bigg) \\
    & \qquad - \bigg( z_X \sum_{i=1}^{m} \sum_{j=1}^{n} W_{q,i,j} \bigg) + m~n~z_X z_W \Bigg]
\end{split}
\end{equation}

\subsection{AveragePool2D} \label{appendix:averagepool2d}

\begin{equation}
\begin{split}
    y
    &= \frac{s_X}{m~n} \sum_{i=1}^{m} \sum_{j=1}^{n} (X_{q,i,j} - z_X) \\
    &= \frac{s_X}{m~n} \Bigg[ \bigg( \sum_{i=1}^{m} \sum_{j=1}^{n} X_{q,i,j}  \bigg) - m~n~z_X \Bigg] \\
    &= s_X \Bigg[ \bigg( \frac{1}{m~n} \sum_{i=1}^{m} \sum_{j=1}^{n} X_{q,i,j}  \bigg) - z_X \Bigg] \\
    &= s_y(y_q - z_y)
\end{split}
\end{equation}
where $X_q$ and $y_q$ are the quantized versions of $X$ and $y$, respectively, $s_X$ and $s_y$ are the scales, while $z_X$ and $z_y$ are the zero points.
Therefore:
\begin{equation}
    y_q = z_y + \frac{s_X}{s_y} \Bigg[ \bigg( \frac{1}{m~n} \sum_{i=1}^{m} \sum_{j=1}^{n} X_{q,i,j}  \bigg) - z_X \Bigg]
\end{equation}

\subsection{ReLU Function} \label{appendix:relu}

\begin{equation}
\begin{split}
    y
    &=
    \begin{cases}
        0 & \text{if $s_x(x_q - z_x) < 0$} \\
        s_x(x_q - z_x) & \text{if $s_x(x_q - z_x) \geq 0$} \\
    \end{cases} \\
    &=
    \begin{cases}
        0 & \text{if $x_q < z_x$} \\
        s_x(x_q - z_x) & \text{if $x_q \geq z_x$} \\
    \end{cases} \\
    &= s_y(y_q - z_y)
\end{split}
\end{equation}
where $x_q$ and $y_q$ are the quantized versions of $x$ and $y$, respectively, $s_x$ and $s_y$ are the scales, while $z_x$ and $z_y$ are the zero points.
Therefore:
\begin{equation}
    y_q = 
    \begin{cases}
        z_y & \text{if $x_q < z_x$} \\
        z_y + \frac{s_x}{s_y} (x_q - z_x) & \text{if $x_q \geq z_x$} \\
    \end{cases}
\end{equation}

\subsection{Softmax Function} \label{appendix:softmax}

\begin{equation}
\begin{split}
    y
    &= \frac{e^{s_x(x_{q,i} - z_x)}}{\sum_{j=1}^{n} e^{s_x(x_{q,j} - z_x)}} \\
    &= \frac{e^{s_x x_{q,i}}}{\sum_{j=1}^{n} e^{s_x x_{q,j}}} \\
    &= s_y(y_{q,i} - z_y)
\end{split}
\end{equation}
where $x_q$ and $y_q$ are the quantized versions of $x$ and $y$, respectively, $s_x$ and $s_y$ are the scales, while $z_x$ and $z_y$ are the zero points.
Therefore:
\begin{equation}
    y_{q,i} = z_y + \frac{e^{s_x x_{q,i}}}{s_y \sum_{j=1}^{n} e^{s_x x_{q,j}}}
\end{equation}

\bibliographystyle{unsrt}  
\bibliography{references}  

\begin{thebibliography}{10}

\bibitem{tinyml-book}
Pete Warden and Daniel Situnayake.
\newblock {\em TinyML}.
\newblock O'Reilly Media, Inc., 12 2019.

\bibitem{dutta2021tinyml}
Lachit Dutta and Swapna Bharali.
\newblock {Tinyml Meets IoT: A Comprehensive Survey}.
\newblock {\em Internet of Things}, 16:100461, 2021.

\bibitem{978-1-68038-141-2}
Microcontroller market size, share \& trends analysis report by product (8-bit, 16-bit, 32-bit), by application (consumer electronics \& telecom, automotive, industrial, medical devices, aerospace \& defense), by region, and segment forecasts, 2023 - 2030.
\newblock Technical Report 978-1-68038-141-2, Grand View Research, 2023.

\bibitem{2022-ai-index-report}
Daniel Zhang, Nestor Maslej, Erik Brynjolfsson, John Etchemendy, Terah Lyons, James Manyika, Helen Ngo, Juan~Carlos Niebles, Michael Sellitto, Ellie Sakhaee, Yoav Shoham, Jack Clark, and Raymond Perrault.
\newblock The ai index 2022 annual report.
\newblock Technical report, AI Index Steering Committee, Stanford Institute for Human-Centered AI, Stanford University, 3 2022.

\bibitem{tinyml-comes-to-embedded-world-2023}
Sally Ward-Foxton.
\newblock Tinyml comes to embedded world 2023.
\newblock {\em EE Times}, 2023.

\bibitem{9166461}
Ramon Sanchez-Iborra and Antonio~F. Skarmeta.
\newblock Tinyml-enabled frugal smart objects: Challenges and opportunities.
\newblock {\em IEEE Circuits and Systems Magazine}, 20(3):4--18, 2020.

\bibitem{9682107}
Samson~Otieno Ooko, Marvin Muyonga~Ogore, Jimmy Nsenga, and Marco Zennaro.
\newblock Tinyml in africa: Opportunities and challenges.
\newblock In {\em 2021 IEEE Globecom Workshops (GC Wkshps)}, pages 1--6, 2021.

\bibitem{MLSYS2021_d2ddea18}
R.~David, J.~Duke, A~Jain, V.~J. Reddi, N.~Jeffries, J.~Li, N.~Kreeger, I.~Nappier, M.~Natraj, S.~Regev, R.~Rhodes, T.~Wang, and P.~Warden.
\newblock Tensorflow lite micro: Embedded machine learning for tinyml systems.
\newblock In A.~Smola, A.~Dimakis, and I.~Stoica, editors, {\em Proceedings of Machine Learning and Systems}, volume~3, pages 800--811, 2021.

\bibitem{WhiteHouse2024}
{The White House}.
\newblock Back to the building blocks: A path toward secure and measurable software.
\newblock Technical report, The White House, 2024.

\bibitem{lin2023embedded}
Hsiao-Ying Lin.
\newblock Embedded artificial intelligence: intelligence on devices.
\newblock {\em Computer}, 56(9):90--93, 2023.

\bibitem{6854370}
Guoguo Chen, Carolina Parada, and Georg Heigold.
\newblock Small-footprint keyword spotting using deep neural networks.
\newblock In {\em 2014 IEEE International Conference on Acoustics, Speech and Signal Processing (ICASSP)}, pages 4087--4091, 2014.

\bibitem{labrador2013human}
Miguel~A Labrador and Oscar D~Lara Yejas.
\newblock {\em Human activity recognition: Using wearable sensors and smartphones}.
\newblock CRC Press, 2013.

\bibitem{9463524}
Alexander Wong, Mahmoud Famuori, Mohammad~Javad Shafiee, Francis Li, Brendan Chwyl, and Jonathan Chung.
\newblock Yolo nano: a highly compact you only look once convolutional neural network for object detection.
\newblock In {\em 2019 Fifth Workshop on Energy Efficient Machine Learning and Cognitive Computing - NeurIPS Edition (EMC2-NIPS)}, pages 22--25, 2019.

\bibitem{9837510}
Anargyros Gkogkidis, Vasileios Tsoukas, Stefanos Papafotikas, Eleni Boumpa, and Athanasios Kakarountas.
\newblock A tinyml-based system for gas leakage detection.
\newblock In {\em 2022 11th International Conference on Modern Circuits and Systems Technologies (MOCAST)}, pages 1--5, 2022.

\bibitem{9700573}
Maria~Francesca Alati, Giancarlo Fortino, Juan Morales, Jose~M. Cecilia, and Pietro Manzoni.
\newblock Time series analysis for temperature forecasting using tinyml.
\newblock In {\em 2022 IEEE 19th Annual Consumer Communications \& Networking Conference (CCNC)}, pages 691--694, 2022.

\bibitem{justus2018predicting}
Daniel Justus, John Brennan, Stephen Bonner, and Andrew~Stephen McGough.
\newblock {Predicting the Computational Cost of Deep Learning Models}.
\newblock In {\em IEEE International Conference on Big Data}, pages 3873--3882. IEEE, 2018.

\bibitem{lin2022device}
Ji~Lin, Ligeng Zhu, Wei-Ming Chen, Wei-Chen Wang, Chuang Gan, and Song Han.
\newblock On-device training under 256kb memory.
\newblock {\em Advances in Neural Information Processing Systems}, 35:22941--22954, 2022.

\bibitem{pavan2024tybox}
Massimo Pavan, Eugeniu Ostrovan, Armando Caltabiano, and Manuel Roveri.
\newblock Tybox: An automatic design and code generation toolbox for tinyml incremental on-device learning.
\newblock {\em ACM Transactions on Embedded Computing Systems}, 23(3):1--27, 2024.

\bibitem{ravaglia2021tinyml}
Leonardo Ravaglia, Manuele Rusci, Davide Nadalini, Alessandro Capotondi, Francesco Conti, and Luca Benini.
\newblock A tinyml platform for on-device continual learning with quantized latent replays.
\newblock {\em IEEE Journal on Emerging and Selected Topics in Circuits and Systems}, 11(4):789--802, 2021.

\bibitem{pasti2024latent}
Francesco Pasti, Marina Ceccon, Davide~Dalle Pezze, Francesco Paissan, Elisabetta Farella, Gian~Antonio Susto, and Nicola Bellotto.
\newblock Latent distillation for continual object detection at the edge.
\newblock {\em Workshop on Computational Aspects of Deep Learning, European Conference on Computer Vision (ECCV)}, 2024.

\bibitem{jacob2018quantization}
Benoit Jacob, Skirmantas Kligys, Bo~Chen, Menglong Zhu, Matthew Tang, Andrew Howard, Hartwig Adam, and Dmitry Kalenichenko.
\newblock Quantization and training of neural networks for efficient integer-arithmetic-only inference.
\newblock In {\em Proceedings of the IEEE conference on computer vision and pattern recognition}, pages 2704--2713, 2018.

\bibitem{han2016deepcompressioncompressingdeep}
Song Han, Huizi Mao, and William~J. Dally.
\newblock {Deep Compression: Compressing Deep Neural Networks with Pruning, Trained Quantization and Huffman Coding}.
\newblock In {\em International Conference on Learning Representations}, 2016.

\bibitem{hinton2015distilling}
Geoffrey Hinton.
\newblock Distilling the knowledge in a neural network.
\newblock {\em NeurIPS Workshops}, 2014.

\bibitem{howard2017mobilenets}
Andrew~G. Howard, Menglong Zhu, Bo~Chen, Dmitry Kalenichenko, Weijun Wang, Tobias Weyand, Marco Andreetto, and Hartwig Adam.
\newblock Mobilenets: Efficient convolutional neural networks for mobile vision applications, 2017.

\bibitem{iandola2016squeezenet}
Forrest~N Iandola.
\newblock Squeezenet: Alexnet-level accuracy with 50x fewer parameters and< 0.5 mb model size.
\newblock {\em arXiv preprint arXiv:1602.07360}, 2016.

\bibitem{8578384}
Benoit Jacob, Skirmantas Kligys, Bo~Chen, Menglong Zhu, Matthew Tang, Andrew Howard, Hartwig Adam, and Dmitry Kalenichenko.
\newblock Quantization and training of neural networks for efficient integer-arithmetic-only inference.
\newblock In {\em 2018 IEEE/CVF Conference on Computer Vision and Pattern Recognition}, pages 2704--2713, 2018.

\bibitem{alajlan2022tinyml}
Norah~N Alajlan and Dina~M Ibrahim.
\newblock {TinyML: Enabling of Inference Deep Learning Models on Ultra-Low-Power IoT Edge Devices for AI Applications}.
\newblock {\em Micromachines}, 13(6):851, 2022.

\bibitem{msrc-2019-bluehat}
Matt Miller.
\newblock Trends, challenges, and strategic shifts in the software vulnerability mitigation landscape.
\newblock \url{https://github.com/microsoft/MSRC-Security-Research/blob/master/presentations/2019_02_BlueHatIL/}, 2019.
\newblock {Microsoft Security Response Center (MSRC)}.

\bibitem{chromium-2020-memory-safety}
{The Chromium Projects}.
\newblock Memory safety.
\newblock \url{https://www.chromium.org/Home/chromium-security/memory-safety}, 2020.

\bibitem{grgic2018comparison}
H~Grgic, Branko Mihaljevi{\'c}, and Aleksander Radovan.
\newblock {Comparison of Garbage Collectors in Java Programming Language}.
\newblock In {\em International Convention on Information and Communication Technology, Electronics and Microelectronics (MIPRO)}, pages 1539--1544. IEEE, 2018.

\bibitem{klabnik2023rust}
Steve Klabnik and Carol Nichols.
\newblock {\em {The Rust programming language}}.
\newblock No Starch Press, 2023.

\bibitem{rust-in-android-platform}
Jeff Vander~Stoep and Stephen Hines.
\newblock Rust in the android platform.
\newblock \url{https://security.googleblog.com/2021/04/rust-in-android-platform.html}, 2021.
\newblock {Android Security}.

\bibitem{lkml-linux-6.1}
Linus Torvalds.
\newblock Linux 6.1.
\newblock \url{https://lkml.org/lkml/2022/12/11/206}, 2022.
\newblock {Linux Kernel Mailing List Archive}.

\bibitem{jung2021safe}
Ralf Jung, Jacques-Henri Jourdan, Robbert Krebbers, and Derek Dreyer.
\newblock Safe systems programming in rust.
\newblock {\em Communications of the ACM}, 64(4):144--152, 2021.

\bibitem{plauska2022performance}
Ignas Plauska, Agnius Liutkevi{\v{c}}ius, and Audron{\.e} Janavi{\v{c}}i{\=u}t{\.e}.
\newblock Performance evaluation of c/c++, micropython, rust and tinygo programming languages on esp32 microcontroller.
\newblock {\em Electronics}, 12(1):143, 2022.

\bibitem{borgsmuller2021rust}
Nico Borgsm{\"u}ller.
\newblock {\em The Rust programming language for embedded software development}.
\newblock PhD thesis, Technische Hochschule Ingolstadt, 2021.

\bibitem{the-rust-programming-language}
Steve Klabnik and Carol Nichols.
\newblock {\em The Rust Programming Language}.
\newblock No Starch Press, 2023.

\bibitem{MLSYS2021_c4d41d96}
Colby Banbury, Chuteng Zhou, Igor Fedorov, Ramon Matas, Urmish Thakker, Dibakar Gope, Vijay Janapa~Reddi, Matthew Mattina, and Paul Whatmough.
\newblock Micronets: Neural network architectures for deploying tinyml applications on commodity microcontrollers.
\newblock In A.~Smola, A.~Dimakis, and I.~Stoica, editors, {\em Proceedings of Machine Learning and Systems}, volume~3, pages 517--532, 2021.

\bibitem{courbariaux2016binarized}
Matthieu Courbariaux, Itay Hubara, Daniel Soudry, Ran El-Yaniv, and Yoshua Bengio.
\newblock Binarized neural networks: Training deep neural networks with weights and activations constrained to +1 or -1, 2016.

\bibitem{TheRustReference2024}
{The Rust Reference}.
\newblock \url{https://doc.rust-lang.org/reference/procedural-macros.html#attribute-macros}.
\newblock {The Rust Foundation, August 2024}.

\bibitem{ATmega328}
Microchip Technology Inc.
\newblock {\em ATmega48A/PA/88A/PA/168A/PA/328/P Data Sheet}, September 2020.
\newblock Rev. B.

\bibitem{TFLiteMemOpt}
Juhyun Lee and Yury Pisarchyk.
\newblock {Optimizing TensorFlow Lite Runtime Memory}.
\newblock \url{https://blog.tensorflow.org/2020/10/optimizing-tensorflow-lite-runtime.html}.
\newblock {Google, August 2024}.

\bibitem{EdgeImpulseMemOpt}
Louis Moreau.
\newblock {Introducing EON Compiler (RAM optimized)}.
\newblock \url{https://www.edgeimpulse.com/blog/introducing-eon-compiler-ram-optimized/}.
\newblock {Edge Impulse, August 2024}.

\bibitem{liang2021pruning}
Tailin Liang, John Glossner, Lei Wang, Shaobo Shi, and Xiaotong Zhang.
\newblock Pruning and quantization for deep neural network acceleration: A survey.
\newblock {\em Neurocomputing}, 461:370--403, 2021.

\bibitem{Zhang2022}
Y.~Zhang, D.~Wijerathne, Z.~Li, and T.~Mitra.
\newblock {Power-Performance Characterization of TinyML Systems}.
\newblock In {\em Proc. of IEEE Int. Conf. on Computer Design (ICCD)}, pages 644--651, 2022.

\bibitem{TFLMSine2024}
TensorFlow.
\newblock Hello world example.
\newblock \url{https://github.com/tensorflow/tflite-micro/tree/main/tensorflow/lite/micro/examples/hello_world}.
\newblock {Google, August 2024}.

\bibitem{TFLMSpeech2024}
TensorFlow.
\newblock Micro speech example.
\newblock \url{https://github.com/tensorflow/tflite-micro/tree/main/tensorflow/lite/micro/examples/micro_speech}.
\newblock {Google, August 2024}.

\bibitem{TFLMPerson2024}
TensorFlow.
\newblock Person detection example.
\newblock \url{https://github.com/tensorflow/tflite-micro/tree/main/tensorflow/lite/micro/examples/person_detection}.
\newblock {Google, August 2024}.

\bibitem{TinyConv}
Jisu Kwon and Daejin Park.
\newblock Toward data-adaptable tinyml using model partial replacement for resource frugal edge device.
\newblock In {\em The International Conference on High Performance Computing in Asia-Pacific Region}, page 133–135. ACM, 2021.

\bibitem{warden2018speech}
Pete Warden.
\newblock Speech commands: A dataset for limited-vocabulary speech recognition, 2018.

\bibitem{chowdhery2019visual}
Aakanksha Chowdhery, Pete Warden, Jonathon Shlens, Andrew Howard, and Rocky Rhodes.
\newblock Visual wake words dataset, 2019.

\bibitem{ESP32FP}
Espressif.
\newblock Report bugs.
\newblock \url{https://www.esp32.com/viewtopic.php?t=800}.
\newblock {August 2024}.

\bibitem{Chen2020}
Tse-Wei Chen, Wei Tao, Deyu Wang, Dongchao Wen, Kinya Osa, and Masami Kato.
\newblock Hardware architecture of embedded inference accelerator and analysis of algorithms for depthwise and large-kernel convolutions.
\newblock In Adrien Bartoli and Andrea Fusiello, editors, {\em Computer Vision -- ECCV 2020 Workshops}, pages 3--17, Cham, 2020. Springer International Publishing.

\bibitem{Hacene2020}
Ghouthi~Boukli Hacene, Vincent Gripon, Matthieu Arzel, Nicolas Farrugia, and Yoshua Bengio.
\newblock Quantized guided pruning for efficient hardware implementations of deep neural networks.
\newblock In {\em 2020 18th IEEE International New Circuits and Systems Conference (NEWCAS)}, pages 206--209, 2020.

\bibitem{Reddi2020}
Vijay~Janapa Reddi, Christine Cheng, David Kanter, Peter Mattson, Guenther Schmuelling, Carole-Jean Wu, Brian Anderson, Maximilien Breughe, Mark Charlebois, William Chou, Ramesh Chukka, Cody Coleman, Sam Davis, Pan Deng, Greg Diamos, Jared Duke, Dave Fick, J.~Scott Gardner, Itay Hubara, Sachin Idgunji, Thomas~B. Jablin, Jeff Jiao, Tom~St. John, Pankaj Kanwar, David Lee, Jeffery Liao, Anton Lokhmotov, Francisco Massa, Peng Meng, Paulius Micikevicius, Colin Osborne, Gennady Pekhimenko, Arun Tejusve~Raghunath Rajan, Dilip Sequeira, Ashish Sirasao, Fei Sun, Hanlin Tang, Michael Thomson, Frank Wei, Ephrem Wu, Lingjie Xu, Koichi Yamada, Bing Yu, George Yuan, Aaron Zhong, Peizhao Zhang, and Yuchen Zhou.
\newblock {MLPerf Inference Benchmark}.
\newblock In {\em 2020 ACM/IEEE 47th Annual International Symposium on Computer Architecture (ISCA)}, pages 446--459, 2020.

\bibitem{inference-tiny-12}
{MLPerf v1.2 Results Inference Tiny}.
\newblock \url{https://mlcommons.org/benchmarks/inference-tiny/}, 2024.

\bibitem{tinyml-continual-2021}
Leonardo Ravaglia, Manuele Rusci, Davide Nadalini, Alessandro Capotondi, Francesco Conti, and Luca Benini.
\newblock A tinyml platform for on-device continual learning with quantized latent replays.
\newblock {\em IEEE Journal on Emerging and Selected Topics in Circuits and Systems}, 11(4):789--802, 2021.

\bibitem{tinyml-federated-2024}
M.~Ficco, A.~Guerriero, E.~Milite, F.~Palmieri, R.~Pietrantuono, and S.~Russo.
\newblock Federated learning for iot devices: Enhancing tinyml with on-board training.
\newblock {\em Information Fusion}, 104:102189, 2024.

\end{thebibliography}

\end{document}